\documentclass[USenglish,oneside,twocolumn]{article}

\usepackage[utf8]{inputenc}

\usepackage[big]{dgruyter_NEW}

\usepackage{amsmath,amsfonts,bm}

\def\eqref#1{equation~\ref{#1}}

\def\1{\bm{1}}

\def\eps{{\varepsilon}}

\def\rl{{\textnormal{l}}}

\DeclareMathAlphabet{\mathsfit}{\encodingdefault}{\sfdefault}{m}{sl}
\SetMathAlphabet{\mathsfit}{bold}{\encodingdefault}{\sfdefault}{bx}{n}

\usepackage{amsmath,amssymb,amsfonts}
\usepackage{algorithmic}
\usepackage{graphicx}
\usepackage{textcomp}
\usepackage{xcolor}
\usepackage{subfig}
\usepackage{endnotes}

\makeatletter
\def\plist@algorithm{\algorithmname\space}
\makeatother

\usepackage{hyperref}
\usepackage{url}
\usepackage[ruled]{algorithm}
\usepackage{amsthm}
\usepackage{multirow}
\usepackage{multicol}
\usepackage{xspace}
\usepackage{bm}
\usepackage{comment}
\usepackage{booktabs}
\usepackage[numbers,sort&compress]{natbib}
\usepackage{babel}
\usepackage[toc,page]{appendix}

\newcommand{\MF}{\mathcal{F}}

\newcommand{\btheta}{\theta}
\newcommand{\calL}{\ensuremath{\mathcal {L}}}

\newcommand{\g}{\ensuremath{\mathbf g}}
\newcommand{\Id}{\ensuremath{\mathbf I}}

\newcommand{\loss}{\mathcal{L}}
\newcommand{\name}{\textsc{DPlis}\xspace}

\newtheorem{myDef}{Definition} 
\newtheorem{myTheo}{Theorem}
\newtheorem{myLe}{Lemma}
\newtheorem{myCor}{Corollary}

\newcommand{\para}[1]{\noindent\textbf{#1.}\xspace}

\newcommand{\myendnote}[2]{\stepcounter{endnote}\endnotetext{#1\label{#2}}\footref{#2}}
\newcommand{\myendref}[1]{\footref{#1}}
 
\DOI{foobar}

\cclogo{\includegraphics{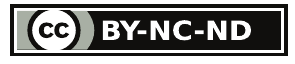}}
  
\begin{document}
 
   \author[1]{Wenxiao Wang}

   \author[2]{Tianhao Wang}

   \author[3]{Lun Wang}

   \author[4]{Nanqing Luo}
   
   \author[5]{Pan Zhou}
   
   \author[6]{Dawn Song}
   
   \author[7]{Ruoxi Jia}

   \affil[1]{Tsinghua University. \mbox{E-mail: wangwx20@mails.tsinghua.edu.cn}}

   \affil[2]{Harvard University. \mbox{E-mail: tianhaowang@fas.harvard.edu}}

   \affil[3]{University of California, Berkeley. \mbox{E-mail: wanglun@berkeley.edu}}

   \affil[4]{Huazhong University of Science and Technology. \mbox{E-mail: u201714868@hust.edu.cn}}
   
   \affil[5]{Huazhong University of Science and Technology. \mbox{E-mail: panzhou@hust.edu.cn}}
   
   \affil[6]{University of California, Berkeley. \mbox{E-mail: dawnsong@cs.berkeley.edu}}
   
   \affil[7]{Virginia Tech. \mbox{E-mail: ruoxijia@vt.edu}}

  \title{\huge \name: Boosting Utility of Differentially Private Deep Learning via Randomized Smoothing}

  \runningtitle{\name: Boosting Utility of Differentially Private Deep Learning via Randomized Smoothing}


  \begin{abstract}
{Deep learning techniques have achieved remarkable performance in wide-ranging tasks. However, when trained on privacy-sensitive datasets, the model parameters may expose private information in training data. Prior attempts for differentially private training, although offering rigorous privacy guarantees, lead to much lower model performance than the non-private ones. Besides, different runs of the same training algorithm produce models with large performance variance. To address these issues, we propose \name--Differentially Private Learning wIth Smoothing. The core idea of \name is to construct a smooth loss function that favors noise-resilient models lying in large flat regions of the loss landscape. We provide theoretical justification for the utility improvements of \name. Extensive experiments also demonstrate that \name can effectively boost model quality and training stability under a given privacy budget.}
\end{abstract}

  \journalname{Proceedings on Privacy Enhancing Technologies}
\DOI{Editor to enter DOI}
  \startpage{1}
  \received{..}
  \revised{..}
  \accepted{..}

  \journalyear{..}
  \journalvolume{..}
  \journalissue{..}

\maketitle

\section{Introduction}

Recent advances in deep neural networks (DNNs) has led to state-of-the-art performances in a wide variety of tasks, including, among others, image recognition and natural language processing. The availability of large datasets is indispensable for these advances. However, in many application domains of deep learning, such as healthcare, finance, and location-based services, the data may contain privacy-sensitive information. As DNNs tend to memorize training data~\cite{fredrikson2015model,carlini2019secret,zhang2017efficient}, the sensitive information might be leaked with the release of the model~\cite{shokri2017membership,salem2018ml,rahman2018membership}.

Differential privacy (DP), a canonical privacy notion that provides provable privacy guarantees, has been applied to mitigate training data leakage. DP aims at hiding the presence of every individual record from the output of an algorithm performed on private data. To achieve this, DP characterizes the sensitivity of output to the change of one record in an arbitrary input dataset and further adds noise to the output that is proportional to the sensitivity. The characterization of sensitivity is challenging for DNNs, as the dependency of the large size parameters on the training data is difficult to understand and trace. 

Prior work presents two general threads of ideas to deal with the challenge of sensitivity characterization in deep learning. One is to recognize that DNNs are mostly trained iteratively via stochastic gradient descent (SGD) methods and thus one can bound the sensitivity of each SGD step by clipping gradient norm and then perturb the gradients accordingly. By ensuring that each SGD step is differentially private, the final output model satisfies a certain level of DP given the composition theorem. As training DNNs involves a large number of iterations, it is necessary to tightly track cumulative privacy loss during training and halt when the loss hits the privacy budget. Moments accountant~\cite{DBLP:conf/ccs/AbadiCGMMT016,wang2019subsampled} provides a much tighter composition analysis for Gaussian noise applied to subsampled data, compared to the standard advanced composition theorem~\cite{dwork2014algorithmic} in DP. The combination of the noisy SGD and moment accountants is often referred to as DP-SGD and has been widely used for building differentially private DNNs. However, despite DP-SGD providing a principled, easily implementable framework for building differentially private DNNs, the \emph{prediction accuracy} of the privately trained models is severely impaired due to the large noise added into each SGD step. Another side-effect of DP-SGD often overlooked by the previous work is the lack of \emph{stability}--different runs of the same algorithm results in drastically different models in terms of prediction accuracy.

Another thread of ideas to overcome the difficulty of sensitivity characterization in DNNs is based on Private Aggregation of Teacher Ensembles (PATE)~\cite{papernot2016semi,papernot2018scalable}. PATE first trains an ensemble of models (i.e., teacher models) on private data and then aggregates the predictions from different teacher models to label public data in a differentially private manner. Since post-processing differentially private results will not affect the privacy guarantee, a differentially private model can be directly constructed via training on the labeled public data. Overall, PATE converts the problem of characterizing the sensitivity of a training algorithm to that of a label aggregation function, which is much easier to analyze. However, PATE requires access to a large public dataset similar to private data and this prerequisite cannot always be satisfied in practice.

Hence, in this paper, we focus on DP-SGD and aim at boosting the utility of DP-SGD in terms of both prediction accuracy and stability. Existing improvement strategies for DP-SGD are based on the idea of modifying the design of \emph{optimization algorithms} and \emph{model architectures}, such as post-processing the noisy gradients in a way that can reduce noise variance~\cite{wang2019dplssgd}, adjusting the gradient clipping threshold to the norms of the SGD updates~\cite{thakkar2019differentially}, adopting a different error backpropagation method~\cite{lee2020differentially}, and using a family of bounded activation functions~\cite{TemperedSigmoid}. In this paper, we propose an improvement strategy that is complementary to existing ones. The proposed strategy is motivated by the following question: Can we modify the \emph{optimization objective} to make it more suitable for differentially private learning?

\begin{figure}[t!]
	\centering
	\subfloat[Standard]{
	\includegraphics[width=.24\textwidth]{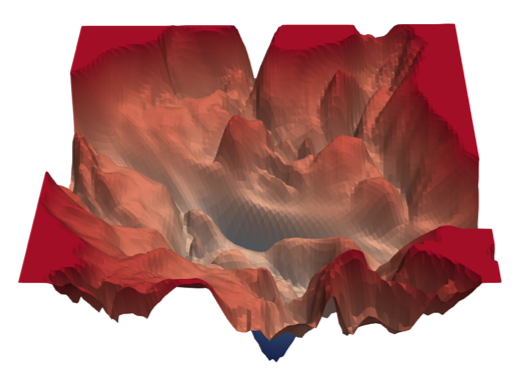}
	\label{fig:intro-sharp}
    }
    \subfloat[Flat]{
	\includegraphics[width=.24\textwidth]{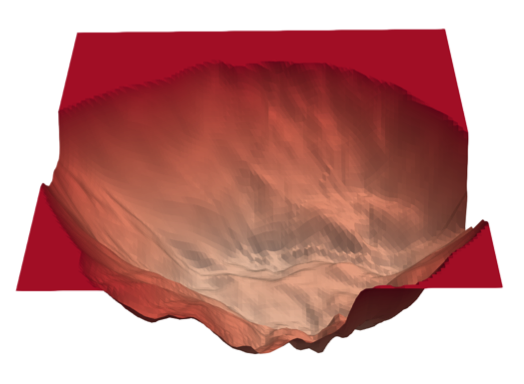}
	\label{fig:intro-flat}
	}
	\caption{Loss surfaces of a ResNet-56 on 1\% training data of CIFAR-10, without and with smoothing. The standard loss surface is irregular and contains many local minima, which is unfavorable given noise injected by DP-SGD. In comparison, a flat loss surface will be more noise-tolerant.} 
	\label{fig:intro-sharp-flat}
\end{figure}

As the first step to answer the question, we examine the issues associated with standard optimization objective functions (i.e., learning loss functions) for DNNs. Figure \ref{fig:intro-sharp} exemplifies the loss surface of a DNN, which is irregular and contains a lot of sharp local minima. Intuitively, such a loss landscape can have detrimental effects on the performance and stability of DP-SGD. Firstly, the noise injected by DP-SGD can result in significant loss increases near each sharp local minimum, which in turn leads to low prediction accuracy. Secondly, with this irregular landscape, noise perturbation along different directions could cause very different loss changes; thus, different runs of the same DP-SGD algorithm might still have large performance variance. Overall, the ``bumpy'' and irregular loss landscape associated with DNNs makes standard learning loss functions unsuitable for performing DP-SGD. Instead, loss functions with flat minima as in Figure \ref{fig:intro-flat} would be more desirable for DP-SGD because they are more noise-tolerant. 

\noindent
\textbf{Contributions.} Inspired by the observations and analysis above, we propose \name, which improves the utility of DP-SGD by smoothing the learning loss function. Algorithmically, we leverage randomized smoothing to smooth the loss function, which convolves Gaussian noise with the original learning loss function. We theoretically connect smoothed loss with the stability and generalization of DP-SGD. Our experiments on diverse datasets and models show that \name alone can often achieve $1\sim 6\%$ prediction accuracy improvement on image data and $8\sim 80\%$ perplexity reduction on language data. At the same time, \name leads to more stable performance across different runs. Particularly, \name is compatible with other DP-SGD improvement strategies that focus on modifying the optimization algorithm (e.g., \cite{wang2019dplssgd,thakkar2019differentially,lee2020differentially}). Thus, \name can potentially further boost the utility of differentially private DNNs when combined with other existing improvement strategies. Moreover, we show that the idea of smoothing learning loss can be easily integrated into PATE---a popular differentially private deep learning framework in the presence of public data. Our experimental results demonstrate that smoothing can usually improve the prediction accuracy by $1\sim 4\%$.

As a side note, smoothing has been widely used in the machine learning community to improve the generalization of models~\cite{wen2018smoothout,chaudhari2019entropy}. However, to the best of our knowledge, our work is the first to investigate smoothing in the context of privacy-preserving learning and provide theoretical bases for the utility gains of smoothing in DP-SGD.

\noindent
\textbf{Paper Organization.} In Section \ref{sec:background}, we include background information regarding Differential Privacy, Deep Learning and their intersections. 
In Section \ref{sec:method}, we motivate our idea with toy examples and present the design of our method, \name. 
In Section \ref{sec:theory}, we derive theoretical results regarding the impact of \name on stability and generalization of DP-SGD.
In Section \ref{sec:eval}, we have the effectiveness of \name verified empirically.
We include a discussion of related work in Section \ref{sec:related}.
Finally we conclude our work in Section \ref{sec:conclusion}.

\section{Background}
\label{sec:background}








\subsection{Differential Privacy}
Given a randomized function that takes a dataset or database as its input, DP~\cite{DBLP:conf/icalp/Dwork06,Dwork2009, Dwork2014} aims to hide the difference of output distribution between two neighboring inputs and provides provable privacy protection against adversaries of arbitrary computational power. 
The formal definition of DP is as follows.

\begin{myDef}
      (\textbf{Differential Privacy}) A randomized function $\MF$ gives $(\eps, \delta)$-differential privacy if for all pairs of adjacent datasets $D, D'\in \mathbb{D}$ that differ in at most one record and all $S \subseteq Range(\MF)$,
	\begin{equation*}
		\Pr[\MF(D)\in S] \leq e^{\eps} \Pr[\MF(D') \in S] + \delta
	\end{equation*}
	\label{DP}
\end{myDef}

A common practice to making an arbitrary function differentially private is through noising \cite{DBLP:conf/icalp/Dwork06,Dwork2009,Dwork2014}, which adds noise of a certain form to the output of the function, with the scale of noise proportional to the sensitivity of the function. The formal definition of sensitivity is as follows.

\begin{myDef}
      (\textbf{Sensitivity}) The sensitivity $\Delta$ of a function $f$ is:
      \begin{equation*}
          \Delta(f) = \max_{D\sim D'} \left\| f(D) - f(D') \right\|,
      \end{equation*}
where $D\sim D'$ indicates that $D$ and $D'$ are two adjacent datasets.
	\label{sensitivity}
\end{myDef}

The sensitivity captures the maximum change of the function outputs when an arbitrary input dataset changes by one entry. When we use $\ell_2$-norm to measure the change, the corresponding sensitivity will be referred to as $\ell_2$-sensitivity or $\Delta_2(f)$.

One prevalent way of noising in the context of differentially private deep learning~\cite{DBLP:conf/ccs/AbadiCGMMT016, DBLP:conf/sp/Yu0PGT19,wang2019dplssgd,DBLP:journals/corr/abs-1910-13659} is the Gaussian mechanism, which adds proper Gaussian noise to the function output based on the $\ell_2$-sensitivity.

\begin{myDef}
      (\textbf{Gaussian Mechanism}) The Gaussian mechanism with parameter $\sigma$, when applied to a function $f   : \mathbb{D} \to \mathbb{R}^K$, 
    adds zero-mean Gaussian noise with variance $\sigma^2$ in each of the $K$ dimensions of $\MF$'s output: $f( \cdot ) + \mathcal{N}(0, \sigma^2 I)$.
\end{myDef}

For $\sigma \geq \frac{ \sqrt{2 \ln( \frac{1.25}{ \delta}) }  \Delta_2(f)} {\eps}$ and $\eps\in (0, 1)$, it is proven that the noised function above is $(\eps, \delta)-$differentially private~\cite{Dwork2014}.

\subsection{Deep Learning}
A neural network $f_\theta$ is a composition of $L$ parametric functions referred to as layers. Each layer consists of neurons, each of which provides one dimension of the layer's output. We denote the $l$th layer as $f_{\theta_l}$ and $\theta_l$ are the associated parameters that control the behavior of the layer. $f_{\theta_l}$ takes as its input the output of the previous layer $f_{\theta_{l-1}}$ and applies a nonlinear transformation to compute its own output. Given an input $x$, a neural network $f_\theta$ performs the following computation to predicts its label:
\begin{equation*}
    f_\theta(x) = f_{\theta_L}\circ\cdots \circ f_{\theta_1}(x)
\end{equation*}
where $\theta=[\theta_1,\ldots,\theta_L]$.

The parameter $\theta$ of a neural network is learned via solving an optimization problem. The optimization objective is the learning loss function $\calL(f_\theta, D)$, which is the average of per-sample loss, i.e. 
\begin{equation*}
    \calL(f_\theta, D) = \frac{1}{|D|}\sum_{x\in D} \calL(f_\theta, x)
\end{equation*}
where $D$ is training data. With a slight abuse of notations, hereinafter, we will use $\calL(\theta)$ as a shorthand notation for the average loss function $\calL(f_\theta, D)$, and use $\calL(\theta, x)$ to denote the sample-level loss function $\calL(f_\theta, x)$.

The optimization problem is often solved via stochastic gradient descent (SGD) methods. At each step of SGD, a batch $B\subseteq D$ of samples is drawn from the training dataset $D$ and the gradient of the average loss $\nabla_{\theta} \calL(\theta) = \frac{1}{|D|}\sum_{x\in D} \nabla_{\theta}\calL(\theta, x)$ will then be approximated by $\frac{1}{|B|}\sum_{x\in B} \nabla_{\theta}\calL(\theta, x)$. With such an approximation, the following rule is applied iteratively to update the parameter:
\begin{equation*}
    \theta \leftarrow \theta - \eta \cdot \frac{1}{|B|}\sum_{x\in B} \nabla_{\theta}\calL(\theta, x)
\end{equation*}
where $\eta$ is the learning rate. 

After the training process completes, the performance of the trained model is evaluated on a held-out test dataset. Since the parameter learning is based on training data, the prediction performance of the trained model is usually good on training data. The ability to also perform well on unseen, test data is referred to as generalization. The performance drop from training to test data is referred to as the generalization gap.

\subsection{Differentially Private Stochastic Gradient Descent}
Differentially private SGD (DP-SGD) was originally proposed in~\cite{DBLP:conf/ccs/AbadiCGMMT016}, and still remains the only general backbone for differentially private deep learning that requires no access to additional public data~\cite{DBLP:conf/ccs/AbadiCGMMT016,DBLP:conf/sp/Yu0PGT19,wang2019dplssgd,DBLP:journals/corr/abs-1910-13659}. The pseudocode of DP-SGD is shown in Algorithm \ref{alg:privsgd}.

\begin{algorithm}[tbp]
	\caption{Differentially Private SGD}\label{alg:privsgd}
	\begin{algorithmic}
	\REQUIRE examples $\{x_1,\ldots,x_N\}$, loss function $\calL(\btheta)=\frac{1}{N}\sum_i \calL(\btheta, x_i)$, learning rate $\eta$, noise multiplier $\sigma$, batch size $L$, clipping threshold $C$. 
		\STATE {\bf Initialize} $\btheta_0$ randomly
		\FOR{$t \in [T]$}
		\STATE {Take a random batch $L_t$ with sampling probability $L/N$ in Poisson subsampling}
		\STATE {\bf Compute gradient}
		\STATE {For each $i\in L_t$, compute $\g_t(x_i) \gets \nabla_{\btheta_t} \calL(\btheta_t, x_i)$}		
		\STATE {\bf Clip gradient}
		\STATE {$\bar{\g}_t(x_i) \gets \g_t(x_i) / \max\big(1, \frac{\|\g_t(x_i)\|_2}{C}\big)$}
		\STATE {\bf Add noise}
		\STATE {$\tilde{\g}_t \gets \frac{1}{L}\left( \sum_i \bar{\g}_t(x_i) + \mathcal{N}(0, \sigma^2 C^2 \Id)\right)$}
		\STATE {\bf Descent}
		\STATE { $\btheta_{t+1} \gets \btheta_{t} - \eta \tilde{\g}_t$}
		\ENDFOR
		\STATE {\bf Output} $\btheta_T$ and compute the overall privacy cost $(\eps, \delta)$ using a privacy accounting method.
	\end{algorithmic}
\end{algorithm}

DP-SGD utilizes the Gaussian mechanism to provide DP guarantees. Since the Gaussian mechanism requires an upper-bound on the sensitivity of the data-dependent function, in DP-SGD, the $\ell_2-$norm of per-sample gradients $\g_t(x_i) \gets \nabla_{\btheta_t} \calL(\btheta_t, x_i)$ are firstly clipped:
\begin{equation*}
    \bar{\g}_t(x_i) \gets \g_t(x_i) / \max\big(1, \frac{\|\g_t(x_i)\|_2}{C}\big),
\end{equation*}
where $C$ is the clipping threshold. This clipping step bounds the $\ell_2-$sensitivity of $\sum_{i} \bar{\g}_t(x_i)$ by $C$ and hence the Gaussian noise that scales proportionally to $C$ will achieve DP. After the gradient is noised, it will be used to update the parameters, following the similar update rule to the regular SGD.

Applying the Gaussian mechanism to each gradient update in the way above allows one to obtain the privacy guarantee for each iteration of SGD. To calculate the privacy guarantee corresponding to the overall SGD algorithm, one could apply the advanced composition theorem for DP~\cite{Dwork2014}. 


\begin{myTheo}
\textbf{Advanced Composition}: For all $\eps, \delta, \delta' \geq 0$, the class of $(\eps, \delta)$-differentially private mechanisms satisfies $(\eps', k \delta + \delta')$-differential privacy under $k$-fold adaptive composition for
\begin{equation*}
    \eps' = \sqrt{2  k \ln{\left( 1 / \delta' \right)}} \eps + k \eps (e^{\eps} - 1).
\end{equation*}
\end{myTheo}

However, it has been shown in~\cite{DBLP:conf/ccs/AbadiCGMMT016} that this generic composition theorem does not yield a tight analysis. Recent works~\cite{DBLP:conf/sp/Yu0PGT19,wang2019subsampled,pmlr-v97-zhu19c} have developed techniques that can produce a much tighter estimate of the overall privacy guarantee for SGD under different subsampling methods and noising mechanisms. Particularly, 
autodp\myendnote{https://github.com/yuxiangw/autodp}{note:autodp} \cite{wang2019subsampled,pmlr-v97-zhu19c} provides an analytical characterization of privacy guarantees for composite differentially private mechanisms and an efficient implementation to track the guarantees. It is compatible with Poisson subsampling---a subsampling method that is typically assumed in DP-SGD. Hence, in this work, we will use autodp as a default way to calculate the privacy parameters for SGD. Unlike the original paper of DP-SGD \cite{DBLP:conf/ccs/AbadiCGMMT016} that uses fixed-size batches as an approximation, we follow Poisson subsampling strictly in our evaluation for rigorous privacy guarantees.



\subsection{Private Aggregation of Teacher Ensembles}
PATE is a popular framework for differentially private deep learning when there exists relevant public data. PATE~\cite{papernot2016semi,papernot2018scalable} consists of three key ingredients: teacher models, an aggregation mechanism, and a student model. 

Teacher models are trained independently on \emph{disjoint} subsets of private data.
The ensemble of teacher models is then used to label the public data. An aggregation mechanism is used to aggregate the class predictions produced by each teacher model. Proper noise is injected into the aggregate predictions to ensure that the labeled public data can achieve certain differential privacy guarantees.

Finally, a student model is trained, usually in a semi-supervised fashion, on public data that is at least partially labeled by an aggregation mechanism. This student model is differentially private as it observes only public data and differentially private labels.

Among the several aggregation mechanisms proposed in prior work~\cite{papernot2016semi,papernot2018scalable}, the most advanced one is Confident-GNMax Aggregator. It first tests with Gaussian noise of scale $\sigma_1$ whether the plurality vote of teacher predictions passes a threshold $T$. If the test is passed, it returns the class with the most votes after the Gaussian noising of scale $\sigma_2$. In this way, priority will be given to labeling public data with sufficiently strong consensus among teacher models.


	
	
\section{Proposed Approach} 
\label{sec:method}




\subsection{Motivating Example}

\begin{figure}[!t]
\centering 
\subfloat[The loss landscape of the toy example described in Equation \ref{eq:toy_example}]{
\includegraphics[width=0.45\linewidth]{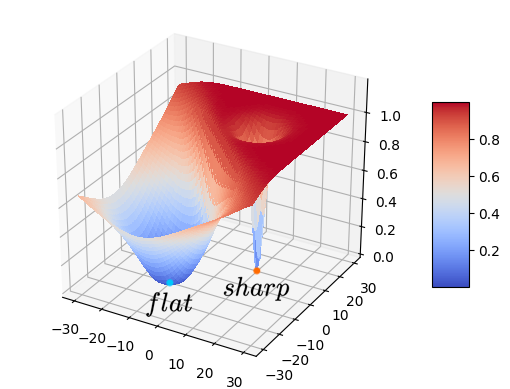} 
\label{fig:toy_loss_surf}
}
\hfil
\subfloat[The smoothed loss landscape of the toy example described in Equation \ref{eq:toy_example}]{
\includegraphics[width=0.45\linewidth]{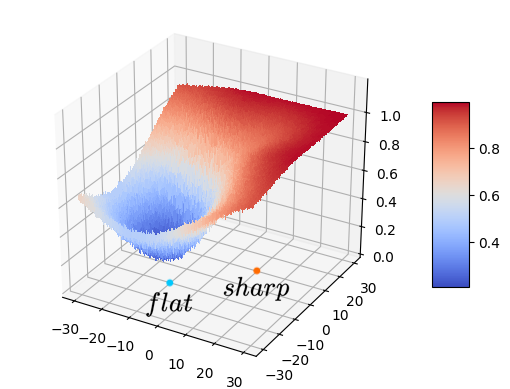} 
\label{fig:toy_smooth_surf}
}

\subfloat[The outputs from 100 independent runs of DP-SGD]{
\includegraphics[width=0.45\linewidth]{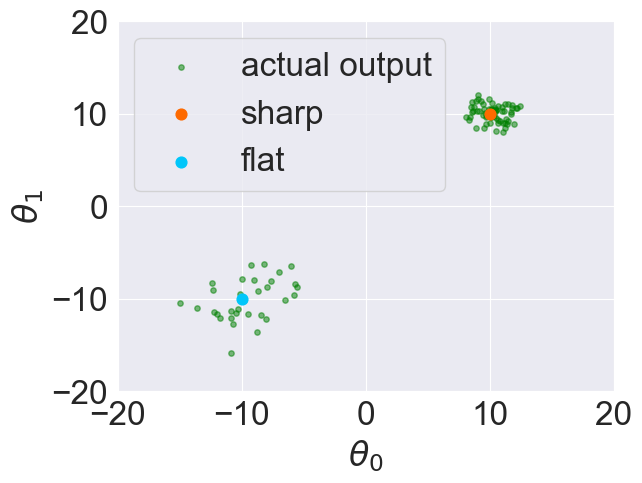} 
\label{fig:toy_sol}
}
\hfil
\subfloat[The outputs from 100 independent runs of DP-SGD with smoothing]{
\includegraphics[width=0.45\linewidth]{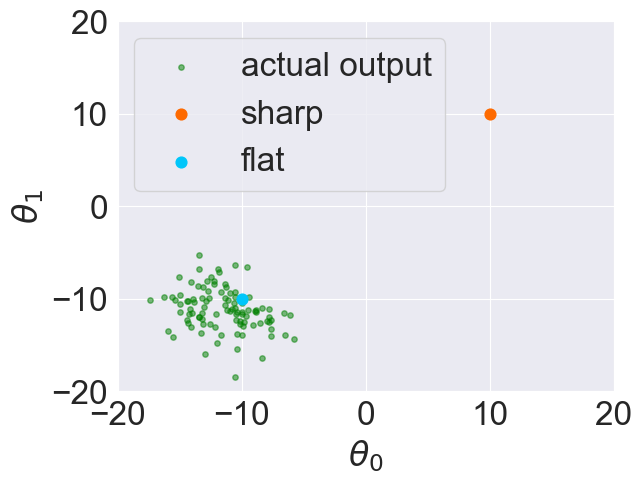} 
\label{fig:toy_smooth_sol}
}

\subfloat[The distribution of loss values from 1000 independent runs of DP-SGD]{
\includegraphics[width=0.45\linewidth]{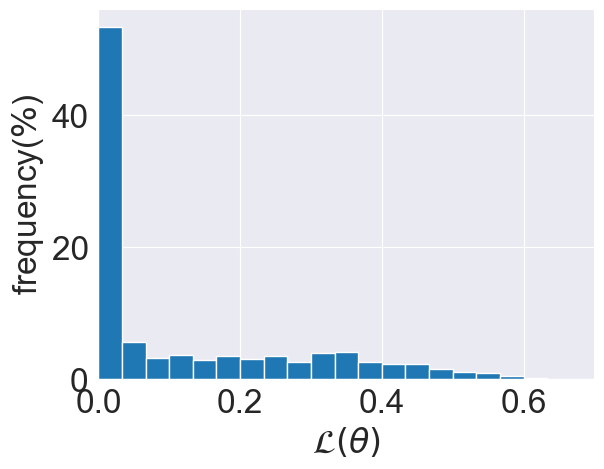} 
\label{fig:toy_loss_hist}
}
\hfil
\subfloat[The distribution of loss values from 1000 independent runs of DP-SGD with smoothing]{
\includegraphics[width=0.45\linewidth]{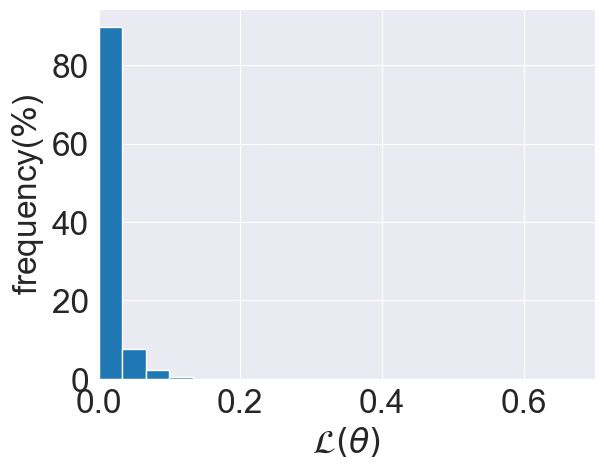} 
\label{fig:toy_smooth_loss_hist}
}

\caption{A toy example illustrating the motivation for \name}
\label{fig:toy_example} 
\end{figure}


Let us first see a simple example to motivate the need for smooth learning loss in differentially private learning. Consider the objective function illustrated in Figure \ref{fig:toy_loss_surf}, 
which is the mixture of two centrosymmetric functions $\MF_u$ and $\MF_v$:
\begin{equation}
\begin{split}
    \min_{\theta \in \mathcal{R}^2} \calL(\theta) 
    & =\MF_u(\theta) \cdot \frac{\mathcal{E}(\theta, u)}{\mathcal{E}(\theta, u) + \mathcal{E}(\theta, v)} \\
    & + \MF_v(\theta) \cdot \frac{\mathcal{E}(\theta, v)}{\mathcal{E}(\theta, u) + \mathcal{E}(\theta, v)}\\
\end{split}
\label{eq:toy_example}
\end{equation}
where $u, v\in \mathcal{R}^2$ are two fixed centers of symmetry, $\MF_u(\theta) = \mathcal{S}\left( \frac{\| \theta - u \|_2}{5} - \frac{5}{\| \theta - u \|_2} \right) $, $\MF_v(\theta) = \mathcal{S}\left( \frac{2\| \theta - v \|_2}{5} - \frac{5}{2\| \theta - v \|_2} \right)  $, the sigmoid function $\mathcal{S}(x) = \frac{1}{1 + e^{-x}}$ and $\mathcal{E}(\theta, x) = e^{- \frac{\| \theta - x \|}{2}}$. 

As shown in Figure~\ref{fig:toy_loss_surf}, 
the loss landscape consists of two local minima with $\calL(\theta) \approx 0$: a flat one and a sharp one. 



Due to the noise injected to protect privacy, the dynamics of DP-SGD will exhibit large variability. In Figure \ref{fig:toy_sol}, we visualize the distribution of $\theta$ obtained from 100 independent runs of DP-SGD and we can see that DP-SGD could converge to the neighborhood of both local minima. 

The noisy nature of DP-SGD degrades not only the performance but the stability of performance, especially when DP-SGD converges to the sharp local minimum. In Figure~\ref{fig:toy_loss_hist}, we present the distribution of loss values from 1000 independent runs of DP-SGD. While loss values are near-zero around both local minima, the loss distribution of actual output models has a long tail. The long tail is mainly caused by the the sharp minimum: a slight change near the sharp minimum could lead to a steep increase of loss. In contrast, a change near the flat minimum would result in a small change of loss.

On the other hand, if we can smooth out the sharp local minima, both the performance and the stability of performance will be greatly improved. Figure~\ref{fig:toy_smooth_surf} illustrates the objective after smoothing. We will dive into the details of the smoothing technique later. With enough smoothing, the sharp local minimum will vanish and only the flat one retains. Running DP-SGD with such a smoothed loss will always converge to the neighborhood of the flatter minimum, which enjoys better performances. Moreover, the variance of loss across different runs of DP-SGD can be mitigated.



\subsection{Main Algorithm}


We leverage the randomized smoothing technique to smooth the learning loss function. Specifically, let $\calL(\theta)$ denote the original loss function. For instance, for classification tasks, a standard choice of $\calL(\theta)$ is cross-entropy. In general, a randomized smoothing technique convolves the density $\mu$ of a random variable with the original loss function:
\begin{align}
    \calL_\text{smooth}(\theta) = \int \calL(\theta')\mu(\theta-\theta')d\theta' =\mathbb{E}_{\Delta \sim \mu}[\calL(\theta+\Delta)]
\end{align}
The intuition underlying such approaches is that convolving two functions yields a new one that is at least as smooth as the smoothest of the two original functions. 

There are various smoothing distributions $\mu$, including Gaussian and uniform distributions on norm balls. We use the Gaussian density function as our default choice of smoothing distribution:
\begin{align}
\label{eqn:l_smooth}
    \calL_\text{smooth}(\theta)& =\mathbb{E}_{\Delta \sim \mathcal{N}(0,\sigma_\text{smooth}^2\Id)}[\calL(\theta+\Delta)] \\
    &= \mathbb{E}_{\theta' \sim \mathcal{N}(\theta,\sigma_\text{smooth}^2\Id)}[\calL(\theta')]
\end{align}
where $\sigma_\text{smooth}$ controls the strength of smoothing: a larger value of $\sigma_\text{smooth}$ indicates a smoother loss function. The choice of Gaussian density function is because we found it is more compatible with DP-SGD which adds Gaussian noise into each gradient update step and as we will see later, its combination with DP-SGD leads to a rigorous generalization bound for the trained model. 

As a simple illustration of randomized smoothing, in Figure~\ref{fig:toy_smooth_surf}, we present the loss landscape of the toy example from Equation~\ref{eq:toy_example} after smoothing. The sharp local minimum vanishes and thus model parameters are effectively guided into the flat local minimum as presented in Figure \ref{fig:toy_smooth_sol}, which results in better and more stable loss values as shown in Figure \ref{fig:toy_smooth_loss_hist}.  We will formally show that the Gaussian density-based randomized smoothing indeed leads to a smoother loss function in the next section.

In addition to smoothing the loss landscape, there is an alternative way of interpreting our proposed loss function $\calL_\text{smooth}(\theta)$. Recall that because of the random noise injection, the output model of DP-SGD is intrinsically random. Essentially, the original learning objective, $\min_\theta \calL(\theta)$, only focuses on optimizing the performance of a single instantiating of the random output model and ignore the fact that the large randomness of the learning process might severely degrade the model performance. Instead, a more reasonable learning objective for DP-SGD is to minimize the expected performance of the output model. Denoting the distribution of model parameters by $\mathcal{H}(\theta)$, we can write out this new learning objective as:
\begin{align}
     \calL_\text{avg}(\theta) = \mathbb{E}_{\theta' \sim \mathcal{H}(\theta)}\calL(\theta')
\end{align}
However, it could be very difficult to analytically characterize $\mathcal{H}(\theta)$, due to the complexity of the model and the learning algorithm. By comparing $\calL_\text{avg}(\theta)$ with $\calL_\text{smooth}(\theta)$, we can see that the gist of our proposed smooth loss is to approximate the output model distribution $\mathcal{H}(\theta)$ via a simple Gaussian distribution. Although this could be a crude approximation, it can lead to easily implementable algorithm and at the same time be mindful of the randomness present in DP-SGD.

Note that the second interpretation of our proposed loss can potentially open up new, improved ways of designing the learning objective for DP-SGD. For instance, one can construct a more accurate approximation to $\mathcal{H}(\theta)$ by leveraging the structure of the learning algorithm. However, in this paper, we will just focus on the smoothness interpretation and present an in-depth investigation of the effect of smoothness on the differentially private learning performance.

When it comes to the implementation of the proposed learning loss, there are a few questions to be addressed: (1) how can we calculate the expectation? and (2) how can we choose the smoothing strength $\sigma_\text{smooth}$? We address the first question by using sample averaging to approximate the expectation. As for setting the smoothing strength, the second interpretation of our smoothed loss can provide us with guidelines. In the second interpretation, $\sigma_\text{smooth}$ captures the spread of the distribution of the output model parameters. Hence, intuitively, $\sigma_\text{smooth}$ should depend on the amount of noise injected by DP-SGD, which further depends on the learning rate, the batch size, and the clipping threshold. Hence, we will factorize $\sigma_\text{smooth}$ into the learning rate $\eta$, the noise multiplier used in DP-SGD $\sigma$, the clipping threshold $C$, as well as a tuning parameter $R$, which we call \emph{smoothing radius}. The smoothing radius captures some complex factors that can impact the spread of parameter distribution, like the training data distribution. 

Formally, the smoothed loss used in the implementation of \name is given as follows:
\begin{equation}
    \min_\theta \frac{1}{K} \sum_{j=1}^K\calL(\btheta + R \cdot \frac{\eta}{L} \cdot  \nu_j) \quad \nu_j \sim\mathcal{N}(0, \sigma^2 C^2 \Id)
\end{equation}
$K$ is the number of samples that we draw from the smoothing distribution. With a larger $K$, the loss above will be a better approximation to $\calL_\text{smooth}(\theta)$ and thus better learning performance could be expected. However, a larger $K$ would at the same makes the learning algorithm more computationally expensive. In general, one can set $K$ to be as large as the computational resource permits. We will present a more detailed study of the effect of $K$ and the smoothing radius in Section~\ref{sec:eval}.

The complete algorithm of DP-SGD with our smoothed loss is provided in Algorithm~\ref{alg:DPSGD_smooth}, where the red texts highlight the difference from the original DP-SGD algorithm.

\begin{algorithm}[tbp]
	\caption{DP-SGD with \name}\label{alg:DPSGD_smooth}
	\begin{algorithmic}
	\REQUIRE examples $\{x_1,\ldots,x_N\}$, loss function $\calL(\btheta)=\frac{1}{N}\sum_i \calL(\btheta, x_i)$, learning rate $\eta$, noise multiplier $\sigma$, batch size $L$, clipping threshold $C$, \textcolor{red}{smoothing radius $R$, number of smoothing samples $K$}. 
		\STATE {\bf Initialize} $\btheta_0$ randomly
		\FOR{$t \in [T]$}
		\STATE {Take a random batch $L_t$ with sampling probability $L/N$ in Poisson subsampling}
		\STATE {\bf Compute gradient}
		\STATE {For each $i\in L_t$, compute \\ \textcolor{red}{$\g_t(x_i) \gets \frac{1}{K} \sum_{j=1}^K\nabla_{\btheta_t} \calL(\btheta_t + R \cdot \frac{\eta}{L} \cdot  \mathcal{N}(0, \sigma^2 C^2 \Id), x_i)$}}	
		\STATE {\bf Clip gradient}
		\STATE {$\bar{\g}_t(x_i) \gets \g_t(x_i) / \max\big(1, \frac{\|\g_t(x_i)\|_2}{C}\big)$}
		\STATE {\bf Add noise}
		\STATE {$\tilde{\g}_t \gets \frac{1}{L}\left( \sum_i \bar{\g}_t(x_i) + \mathcal{N}(0, \sigma^2 C^2 \Id)\right)$}
		\STATE {\bf Descent}
		\STATE { $\btheta_{t+1} \gets \btheta_{t} - \eta \tilde{\g}_t$}
		\ENDFOR
		\STATE {\bf Output} $\btheta_T$ and compute the overall privacy cost $(\eps, \delta)$ using a privacy accounting method.
	\end{algorithmic}
\end{algorithm}

\section{Utility Analysis}
\label{sec:theory}

In this section, we provide the utility analysis for DP-SGD with smoothed loss. We will start by formalizing the smoothness of the loss landscape and further characterize the change in the loss landscape smoothness with randomized smoothing. Moreover, we study the implications of smoothing in improving the stability and generalization of differentially private models. Proofs of lemmas and theorems in this paper can be found in the appendix.
%

\subsection{Smoothness}

A commonly used notion for smoothness is based on the Lipschitz constant of the gradient of a function.

\begin{myDef}[Smoothness]
We say that a function $f$ is $\beta$-smooth if
\begin{align}
    \|\nabla f(x)-\nabla f(y)\|\leq \beta\|x-y\|.
\end{align}
\end{myDef}
$\beta$ is often referred to as the smoothness constant and a smaller $\beta$ indicates a smoother function.

The next lemma characterizes the smoothness constant of a function smoothed with randomized smoothing based on Gaussian density.

\begin{myLe}
Assume that $\mathcal{L}$ is $L$-Lipschitz with respect to the $\ell_2$ norm.
Then, $\mathcal{L}_\text{smooth}$ in Equation (\ref{eqn:l_smooth}) is expected to be $L/\sigma_\text{smooth}$-smooth. In addition, if $\mathcal{L}$ is $\beta$-smooth, then $\mathcal{L}_\text{smooth}$ is expected to be $\min\{L/\sigma_\text{smooth},\beta\}$-smooth.
\label{lem:smooth}
\end{myLe}


Lemma~\ref{lem:smooth} indicates that larger Gaussian noise used in randomized smoothing will lead to a smoother loss landscape. Moreover, note that the first part of Lemma~\ref{lem:smooth} makes no assumption about the smoothness of the original loss function. Hence, even when the original function is unsmooth (i.e., unbounded smooth constant), randomized smoothing can still lead to a smooth function. Indeed, unsmooth loss is quite common in deep learning. For instance, when the ReLU activation function is used, the resulting deep net has unbounded smoothness constant. The second part of the theorem implies that the smoothness constant of $\mathcal{L}_\text{smooth}$ is always less than that of $\mathcal{L}$. In other words, if the original loss function is already smooth, randomized smoothing can only make it smoother.


\subsection{Stability}

With the characterization of smoothness, we can reason about the effect of smoothness on the stability of differentially private learning, i.e., the consistency of model performance across different runs. Particularly, we will use the sum of the expected squared $l_2$-norm of the gradient across all iterations to help measure the stability. Gradients induce the changes in the model parameters and if each gradient has a large variance, then the model parameters will also have a large variance. Hence, the sum of the expected squared $l_2$-norm of the gradient reflects the total variations exhibited during training.




Theorem~\ref{cor:conv} provides a bound on the total expected squared $l_2$-norm of the gradients in terms of the smoothness constant with a simplification through ignoring the effect of gradient clipping. The proof follows from the convergence proof of SGD in~\cite{ghadimi2013stochastic,liu2020distributed}.

\begin{myTheo}
Given an arbitrary $\beta$-smooth learning loss $\mathcal{L}$ and a learning rate $\eta$, assuming $\mathbb{E}[\nabla\mathcal{L}(\theta, x_i)]=\mathbb{E}[\nabla\mathcal{L}(\theta)]$ and $\mathbb{E}[\|\nabla\mathcal{L}(\theta, x_i)-\nabla\mathcal{L}(\theta)\|^2]\leq \sigma^2$ for some parameter $\sigma\leq 0$, we have 
\begin{equation*}
    \frac{1}{T}\sum_{t=1}^T\mathbb{E}[\|\nabla \calL(\theta_t)\|^2] \leq \frac{2\mathbb{E}[\mathcal{L}(\theta_1)] - 2\mathbb{E}[\mathcal{L}(\theta^*)]}{\eta(2-\eta \beta)T} +  \frac{\eta \beta\sigma^2}{(2-\eta \beta)}    
\end{equation*}
\label{cor:conv}
\end{myTheo}


If we use a monotonically decreasing learning rate  $\eta$ and $\eta=\omega(\frac{1}{T})$, then when $T\rightarrow\infty$, the second term on the right-hand side will dominate. 
Since randomized smoothing can effectively reduce the smoothness constant, the left-hand side for the smoothed loss will be less than that for the original loss, which further implies that DP-SGD with \name is more stable than the vanilla DP-SGD.


\subsection{Generalization}
We now switch up to another performance metric for differentially private learning---generalizability of the trained model.

We show with the following theorem that randomized smoothing may help to close generalization gaps. While the same proof idea can be generalized to Poisson subsampling, a simplification is made in the following theorem by considering DP-SGD that uses fixed-size batches instead. This is to derive a more succinct bound that offers more insights.

\begin{myTheo}[Generalization by Smoothness]
Let $\hat{\loss}(\theta)$ be the expected loss on the actual distribution, $\loss(\theta)$ be the corresponding average loss computed on the training data and $\loss(\theta, x)$ be the corresponding loss on sample $x$. 
Let $\loss_\text{train}(\theta, x)$ be the training loss used in DP-SGD on sample $x$, which may differ from $\loss(\theta, x)$.
Assume $\loss(\theta, x)$ is $L$-Lipschitz for every $x$ and $\loss_\text{train}(\theta, x)$ is $\beta$-smooth every $x$, then we have
\begin{align}
\label{eqn:smooth_bound}
    |\mathbb{E}[\hat{\loss}(\theta) - \loss(\theta)] |\leq \frac{2\eta CLT}{n} \cdot (1 + \eta \beta)^{T-1} ,
\end{align}
where $\theta$ is the final parameter from DP-SGD with learning rate $\eta$, clipping threshold $C$, noise multiplier $\sigma$, total number of steps $T$ and total number of training data $n$. The expectations are taken over both the randomness of DP-SGD and the draw of training data.
\label{thm:gap_smooth}
\end{myTheo}

Theorem~\ref{thm:gap_smooth} indicates that the generalization error of a model (i.e., left-hand side of Eqn.~\ref{eqn:smooth_bound}) depends on the smoothness of the loss function used for training the model (i.e., $\beta$ in the right-hand side of Eqn.~\ref{eqn:smooth_bound}) and a smoother training loss can potentially lead to a smaller generalization error. The theorem justifies the advantage of using randomized smoothing during training.

The proof is largely inspired by~\cite{TrainFaster} and is left to the appendix. The key idea of the proof is to leverage a classic result in learning theory that the generalization gap $\mathbb{E}[\hat{\loss}(\theta) - \loss(\theta)]$ can be bounded by $s$ when the learning algorithm is $s$-uniformly stable (i.e. $\mathbb{E}[\loss(\theta, x)]$ may differ at most $s$ when one sample in the training set is replaced). Let $D$ and $D'$ be respectively training set before and after the replacement. Since $\loss(\theta, x)$ is $L$-Lipschitz, the difference $\mathbb{E}[\loss(\theta_D, x)]-\mathbb{E}[\loss(\theta_{D'}, x)]$ is bounded by $L\cdot \mathbb{E}[\|\theta_D - \theta'_D\|]$, where $\theta_D$ and $\theta_{D'}$ represent the models trained on $D$ and $D'$, respectively. Thus, we can analyze $\mathbb{E}[\|\theta_D - \theta'_D\|]$ instead. Fixing the randomness of initial parameters, Gaussian noise, and batch selection, we can analyze how $\theta_D$ and $\theta'_D$ depart from each other through DP-SGD updates using the smoothness of $\loss_\text{train}$. Intuitively, the smoother the training objective is, the less sensitive individual iterations of DP-SGD would be to changes of training points.
At last, we take expectation over the randomness above to obtain the final bound of uniform stability, which is then transferred into the generalization bound.

\section{Evaluation}
\label{sec:eval}

We would like to answer the following five questions with empirical evaluation: 
\begin{enumerate}
    \item How does \name perform compared to vanilla DP-SGD?
    \item Is the performance sensitive to the newly introduced hyper-parameters (i.e., $K$ and smoothing radius)? 
    \item Does randomized smoothing indeed lead to smoother loss landscapes for DNNs?
    \item Does randomized smoothing improve the stability?
    \item Will the gains from smoothing remain combining other improvement strategies?
\end{enumerate}

In addition, we present how the smoothing idea can be extended to improve PATE---a popular framework for differentially private deep learning with public data.

\subsection{Experimental Setup}
\label{sec:evaluationsetup}

\para{Datasets and Models}
To answer question (1), we evaluate \name on various learning tasks.
%
%
We first demonstrate \name's performance improvement on two classic vision benchmarks: \textbf{MNIST} and \textbf{CIFAR-10}.
On these two benchmarks, models are trained entirely from scratch, which we refer to as \emph{non-transfer settings}.
MNIST~\cite{lecun-98} is one of the most commonly used benchmark datasets in deep learning containing 70000 handwritten digit images.
CIFAR-10~\cite{cifar} is another classic benchmark for image classification. 
It consists of 60000 images from 10 different classes with 6000 images each.
We evaluate two different architectures on MNIST and CIFAR-10: multilayer perceptron (MLP) and convolution neural network (CNN).
The MLP is comprised of two hidden layers with 512 (1536) neurons and 128 (128) neurons for MNIST (CIFAR-10) all using ReLU activation. 
The architecture of the CNN is inherited from the official tutorial of tensorflow/privacy\myendnote{https://github.com/tensorflow/privacy}{note:tensorflow_privacy}. 
%




Besides non-transfer settings, given the access to pre-trained feature extractors, we also evaluate \name on more challenging datasets, which we refer to as \emph{transfer settings}. 
Specifically, we evaluate \name in three transfer settings.
\begin{itemize}
    \item \textbf{ImageNet\bm{$\to$}CIFAR-100}: ImageNet~\cite{ILSVRC15} denotes the image classification benchmark in ImageNet Large Scale Visual Recognition Challenge (ILSVRC). It includes a training set with over 1.2 million images and a validation set of 50000 images, all in full resolution, spanning 1000 different classes. CIFAR-100~\cite{cifar} is similar to CIFAR-10 but with 100 different classes. In this setting, we use a ImageNet pre-trained EfficientNet-b0\myendnote{https://github.com/lukemelas/EfficientNet-PyTorch}{note:EfficientNet_PyTorch} \cite{EfficientNet} to extract features. After average-pooling, the extractor provides 1280-dimensional feature vectors, on which an MLP with two 256-neuron hidden layers and ReLU activation is trained to classify CIFAR-100.
    \item \textbf{CelebA\bm{$\to$}PubFig83}: PubFig83\cite{pubfig83} is a dataset of 13838 facial images from 83 public figures with at least 100 images per identity. In our evaluation, we use a pre-processed version of PubFig83\myendnote{https://ic.unicamp.br/\texttildelow chiachia/resources/pubfig83-aligned/}{note:pugfig83_aligned} \cite{pubfig83-aligned}, where images are aligned by the position of the eyes. 50 images from every identity are taken into the testing set with a total of 4150 images, and the remaining 9688 images constitute the training set. CelebA\cite{celeba} is a large-scale dataset of 202599 facial images corresponding to 10177 identities. We remove all 44 identities with only one corresponding image from CelebA, which results in a total of 202555 images and 10133 identities. In this setting, the pre-training is accomplished on this dataset using 10133-way face identification with one image from every identity preserved for validation. We choose ResNet-50\cite{resnet} as the backbone of pre-trained feature extractors, following the implementation of face.evoLVe\myendnote{https://github.com/ZhaoJ9014/face.evoLVe.PyTorch}{note:face_evolve}. After pre-training, average-pooling is applied to the outputs of the last convolution layer to form features with 2048 dimensions, and an MLP with one 32-neuron hidden layer and ReLU activation is trained on such features.
    \item \textbf{CelebA$_{-}$\bm{$\to$}CelebA\bm{$_{1000}$}}: In CelebA$_{-}$\bm{$\to$}CelebA$_{1000}$ setting, 1000 identities with exactly 30 corresponding facial images are picked to form the private dataset, namely CelebA$_{1000}$, while the remaining 9133 identities constitutes the dataset for pre-training, namely CelebA$_{-}$. The pre-training on CelebA$_{-}$ is similar to the one in CelebA\bm{$\to$}PubFig83 setting, with one image from every identity preserved for validation. For the private dataset CelebA$_{1000}$, 25 images from every identity are taken by the training set, and the other 5 images are taken by the testing set, which leads to a training set of size 25000 and a testing set of size 5000. The architecture is the same as CelebA\bm{$\to$}PubFig83 setting.
\end{itemize}
%
%
%



Beyond vision tasks, we also evaluate \name on an NLP task, namely next word prediction on \textbf{Reddit Comments} (Reddit).
Reddit Comments \cite{al2016conversational} is a collection of Reddit posts.
Following the setup of~\cite{kerrigan2020differentially}, we randomly sample 10000 comments as training data and 5000 comments as testing data, and pre-train the model on an additional public data source, Brown corpus~\cite{francis1979brown}, without differential privacy for warm-starting. 
Following \cite{mcmahan2017learning}, we choose to use a pre-selected dictionary from public data instead of to use heavy hitter estimation\cite{qin2016heavy} to generate one from private data. Our dictionary is composed of a total of 36743 words, containing all words with a frequency of at least two from Brown corpus. 
%

%
We evaluate both MLP and LSTM models for the task.
The MLP takes the last 20 tokens as its input and contains three hidden layers with respectively 500, 250, and 50 neurons.
The LSTM model takes a series of words as input and embeds them individually into a learned 256-dimensional space.
The embedded characters are then processed through an LSTM module with 96 nodes. 
Finally, the output of the LSTM module is sent to a fully-connected layer for word predictions. 

\begin{table}[!t]  
\renewcommand{\arraystretch}{1.5} 
\newcommand{\tablewidth}{6} 
\caption{Hyper-parameters} 
\label{tab:hyper-params}

\centering
\resizebox{\linewidth}{!}{ 
\begin{tabular}{c|ccccc}
\hline
\bfseries setting & $L$ & $\frac{\eta}{L}$ & $\eta$ &  $\sigma$ &  $C$ \\ \hline
MNIST & $256$ & $6\times 10^{-4}$ & $0.1536$ & $1.1$ & $1.0$ \\
CIFAR-10 & $256$ & $6\times 10^{-4}$ & $0.1536$ & $1.1$ & $1.0$\\
\multirow{1}{*}{\shortstack{ImageNet\bm{$\to$}CIFAR-100}} & \multirow{1}{*}{$256$} & \multirow{1}{*}{$4\times 10^{-4}$} & \multirow{1}{*}{$0.1024$} &
\multirow{1}{*}{$3.0$}& \multirow{1}{*}{$1.0$} \\
\multirow{1}{*}{\shortstack{CelebA\bm{$\to$}PubFig83}} & \multirow{1}{*}{$256$} & \multirow{1}{*}{$6\times 10^{-3}$} & \multirow{1}{*}{$1.536$} &
\multirow{1}{*}{$2.0$}& \multirow{1}{*}{$1.0$} \\
\multirow{1}{*}{\shortstack{CelebA$_{-}$\bm{$\to$}CelebA$_{1000}$}} & \multirow{1}{*}{$256$} & \multirow{1}{*}{$6\times 10^{-3}$} & \multirow{1}{*}{$1.536$} &
\multirow{1}{*}{$1.3$}& \multirow{1}{*}{$1.0$} \\
Reddit & $64$ & $10^{-3}$ & $0.064$ & $1.1$ & $1.0$\\
\hline
\end{tabular}
}
\end{table}

\para{Training Setups}
For all the reported results, all hyper-parameters for DP-SGD are selected to achieve the best performance of vanilla DP-SGD. \name then sets the exactly same value for the part of hyperparameters that appear in DP-SGD.
The exact choice of hyper-parameters are presented in Table \ref{tab:hyper-params}, where $L$ is (expected) batch size, $\eta$ is learning rate, $\sigma$ is noise multiplier and $C$ is clipping threshold. Batches are sampled through Poisson sampling with probability $\frac{L}{N}$, where $N$ is the size of the training set. 

For \name, unless otherwise specified, we set $K=10$ in both non-transfer and transfer settings and $K=20$ for next word prediction on Reddit. We present results for radius $R=10$ and $R=R_{\text{best}}$, where $R_{\text{best}}$ denotes the radius that achieves the best performance under corresponding privacy budgets. 

In presenting privacy budgets, we fix $\delta$ in $(\eps, \delta)-$differential privacy to be $10^{-5}$ and present only the corresponding $\eps$. Each row in Table~\ref{tab:non-transfer}, Table~\ref{tab:transfer} and Table~\ref{tab:reddit} corresponds to results from a single run.

\subsection{Performance Evaluation}
\label{sec:performance_eval}

\begin{table*}[!t]  
\renewcommand{\arraystretch}{1.3} 
\newcommand{\tablewidth}{7} 
\caption{Evaluation in Non-transfer Settings} 
\label{tab:non-transfer}

\centering
\resizebox{\linewidth}{!}{ 
\begin{tabular}{|c||c||c||l|l|l|l|l}
\hline
\multicolumn{1}{|c||}{\multirow{2}{*}{\bfseries dataset}} & \multirow{2}{*}{\bfseries model} & \multirow{2}{*}{\bfseries smoothing} & \multicolumn{ 4 }{c|}{\bfseries test accuracy} \\ \cline{4-\tablewidth} 
\multicolumn{1}{|c||}{} & & & \multicolumn{1}{c|}{$\eps=1.99$} & \multicolumn{1}{c|}{$\eps=5.01$} & \multicolumn{1}{c|}{$\eps=7.01$} & \multicolumn{1}{c|}{$\eps=10.00$} \\
\hline\hline
\multicolumn{1}{|c||}{\multirow{6}{*}{MNIST}} & \multirow{3}{*}{MLP}& No & $92.60\%$ & $93.92\%$ & $94.11\%$ & $93.35\%$\\ \cline{3-\tablewidth}
\multicolumn{1}{|c||}{}& & $R=10$ & $92.66\%\color{blue}{(+0.06\%)}$ & $94.78\%\color{blue}{(+0.86\%)}$ & $94.84\%\color{blue}{(+0.73\%)}$ & $95.19\%\color{blue}{(+1.84\%)}$ \\ \cline{3-\tablewidth}
\multicolumn{1}{|c||}{}& & $R=R_{\text{best}}$ & $93.58\%\color{blue}{(+0.98\%)}$ & $95.94\%\color{blue}{(+2.02\%)}$ & $96.17\%\color{blue}{(+2.06\%)}$ & $96.00\%\color{blue}{(+2.65\%)}$ \\ \cline{2-\tablewidth}
\multicolumn{1}{|c||}{}& \multirow{3}{*}{CNN} & No & $95.08\%$ & $95.92\%$ & $96.07\%$ & $95.89\%$\\ \cline{3-\tablewidth}
\multicolumn{1}{|c||}{}& & $R=10$ & $96.62\%\color{blue}{(+1.54\%)}$ & $97.49\%\color{blue}{(+1.57\%)}$ & $97.52\%\color{blue}{(+1.45\%)}$ & $98.56\%\color{blue}{(+2.67\%)}$ \\ \cline{3-\tablewidth}
\multicolumn{1}{|c||}{}& & $R=R_{\text{best}}$ & $97.31\%\color{blue}{(+2.23\%)}$ & $98.01\%\color{blue}{(+2.09\%)}$ & $98.48\%\color{blue}{(+2.41\%)}$ & $98.90\%\color{blue}{(+3.01\%)}$ \\ 
\hline\hline
\multicolumn{1}{|c||}{\multirow{6}{*}{CIFAR-10}} & \multirow{3}{*}{MLP} & No & $43.88\%$ & $43.16\%$ & $41.95\%$ & $41.07\%$\\ \cline{3-\tablewidth}
\multicolumn{1}{|c||}{}& & $R=10$ & $44.09\%\color{blue}{(+0.21\%)}$ & $44.48\%\color{blue}{(+1.32\%)}$ & $44.73\%\color{blue}{(+2.78\%)}$ & $44.95\%\color{blue}{(+3.88\%)}$ \\ \cline{3-\tablewidth}
\multicolumn{1}{|c||}{}& & $R=R_{\text{best}}$ & $44.34\%\color{blue}{(+0.46\%)}$ & $46.90\%\color{blue}{(+3.74\%)}$ & $46.92\%\color{blue}{(+4.97\%)}$ & $46.95\%\color{blue}{(+5.88\%)}$ \\ \cline{2-\tablewidth}
\multicolumn{1}{|c||}{}& \multirow{3}{*}{CNN} & No & $49.70\%$ & $58.48\%$ & $60.46\%$ & $61.37\%$\\ \cline{3-\tablewidth}
\multicolumn{1}{|c||}{}& & $R=10$ & $50.09\%\color{blue}{(+0.39\%)}$ & $60.39\%\color{blue}{(+1.91\%)}$ & $62.09\%\color{blue}{(+1.63\%)}$ & $63.22\%\color{blue}{(+1.85\%)}$ \\ \cline{3-\tablewidth}
\multicolumn{1}{|c||}{}& & $R=R_{\text{best}}$ & $50.85\%\color{blue}{(+1.15\%)}$ & $61.75\%\color{blue}{(+3.27\%)}$ & $62.32\%\color{blue}{(+1.86\%)}$ & $64.73\%\color{blue}{(+3.36\%)}$ \\ 
\hline
\end{tabular}
}
\end{table*}

\para{Results in Non-transfer Settings}
As shown in Table \ref{tab:non-transfer}, using \name improves test accuracy of DP-SGD by $0.46\%\sim 5.88\%$ with $R=R_{\text{best}}$ and $0.06\%\sim 3.88\%$ with $R=10$ in non-transfer settings.

An interesting phenomenon is that the test accuracy of vanilla DP-SGD occasionally drops drastically when training proceeds.
As a concrete example, when we train MLP on CIFAR-10, the test accuracy starts to degrade as early as before $\eps=5.01$. 
Though it seems like overfitting, it is actually not since no increasing generalization gap is observed.
We will further discuss this in Section \ref{sec:justification_smoothing}. 
This issue, if not fully addressed, is greatly alleviated by randomized smoothing. 
When training proceeds, no significant drop on test accuracy is observed for \name.


\begin{table*}[!t]  
\renewcommand{\arraystretch}{1.3} 
\newcommand{\tablewidth}{7} 
\caption{Evaluation in Transfer Settings} 
\label{tab:transfer}

\centering
\resizebox{\linewidth}{!}{ 
\begin{tabular}{|c||c||l|l|l|l|l|l}
\hline
\multicolumn{1}{|c||}{\bfseries setting} & \bfseries smoothing & \multicolumn{ 5 }{c|}{\bfseries test accuracy} \\ \hline\hline
\multicolumn{1}{|c||}{\multirow{4}{*}{\shortstack{ImageNet\bm{$\to$}\\CIFAR-100}}} & & \multicolumn{1}{c|}{$\eps=0.39$} & \multicolumn{1}{c|}{$\eps=0.54$} & \multicolumn{1}{c|}{$\eps=0.66$} & \multicolumn{1}{c|}{$\eps=0.77$} & \multicolumn{1}{c|}{$\eps=0.86$} \\ \cline{2-\tablewidth}
\multicolumn{1}{|c||}{}&  No & $34.75\%$ & $42.34\%$ & $44.52\%$ & $46.00\%$ & $46.78\%$\\ \cline{2-\tablewidth}
\multicolumn{1}{|c||}{}&  $R=10$ & $35.73\%\color{blue}{(+0.98\%)}$ & $43.32\%\color{blue}{(+0.98\%)}$ & $46.59\%\color{blue}{(+2.07\%)}$ & $49.02\%\color{blue}{(+3.02\%)}$ & $49.70\%\color{blue}{(+2.92\%)}$ \\ \cline{2-\tablewidth}
\multicolumn{1}{|c||}{}& $R=R_{\text{best}}$ & $35.98\%\color{blue}{(+1.23\%)}$ & $43.76\%\color{blue}{(+1.42\%)}$ & $48.00\%\color{blue}{(+3.48\%)}$ & $51.04\%\color{blue}{(+5.04\%)}$ & $52.04\%\color{blue}{(+5.26\%)}$ \\ 
\hline
\hline
\multicolumn{1}{|c||}{\multirow{4}{*}{\shortstack{CelebA\bm{$\to$}\\PubFig83}}} & & \multicolumn{1}{c|}{$\eps=2.84$} & \multicolumn{1}{c|}{$\eps=4.06$} & \multicolumn{1}{c|}{$\eps=5.04$} & \multicolumn{1}{c|}{$\eps=5.89$} & \multicolumn{1}{c|}{$\eps=6.64$} \\ \cline{2-\tablewidth}
\multicolumn{1}{|c||}{}& No & $41.20\%$ & $53.45\%$ & $55.95\%$ & $59.47\%$ & $62.17\%$\\ \cline{2-\tablewidth}
\multicolumn{1}{|c||}{}& $R=10$ & $41.76\%\color{blue}{(+0.56\%)}$ & $54.53\%\color{blue}{(+1.08\%)}$ & $61.76\%\color{blue}{(+5.81\%)}$& $62.84\%\color{blue}{(+3.37\%)}$ & $64.84\%\color{blue}{(+2.67\%)}$ \\ \cline{2-\tablewidth}
\multicolumn{1}{|c||}{}& $R=R_{\text{best}}$ & $43.16\%\color{blue}{(+1.96\%)}$ & $57.16\%\color{blue}{(+3.71\%)}$ & $61.76\%\color{blue}{(+5.81\%)}$ & $64.27\%\color{blue}{(+4.80\%)}$ & $66.24\%\color{blue}{(+4.07\%)}$ \\ 
\hline\hline
\multicolumn{1}{|c||}{\multirow{4}{*}{\shortstack{CelebA$_{-}$\bm{$\to$}\\CelebA$_{1000}$}}} & & \multicolumn{1}{c|}{$\eps=4.88$} & \multicolumn{1}{c|}{$\eps=6.06$} & \multicolumn{1}{c|}{$\eps=7.12$} & \multicolumn{1}{c|}{$\eps=8.03$} & \multicolumn{1}{c|}{$\eps=8.87$} \\ \cline{2-\tablewidth}
\multicolumn{1}{|c||}{}& No & $19.68\%$ & $25.74\%$ & $31.54\%$ & $34.12\%$ & $36.44\%$\\ \cline{2-\tablewidth}
\multicolumn{1}{|c||}{}& $R=10$ & $21.46\%\color{blue}{(+1.78\%)}$ & $29.86\%\color{blue}{(+4.12\%)}$ & $34.32\%\color{blue}{(+2.78\%)}$& $38.18\%\color{blue}{(+4.06\%)}$ & $40.32\%\color{blue}{(+3.88\%)}$ \\ \cline{2-\tablewidth}
\multicolumn{1}{|c||}{}& $R=R_{\text{best}}$ & $22.54\%\color{blue}{(+2.86\%)}$ & $30.38\%\color{blue}{(+4.64\%)}$ & $35.58\%\color{blue}{(+4.04\%)}$ & $38.88\%\color{blue}{(+4.76\%)}$ & $41.08\%\color{blue}{(+4.64\%)}$ \\
\hline
\end{tabular}
}
\end{table*}

\begin{table}[!b]  
\renewcommand{\arraystretch}{1.5} 
\newcommand{\tablewidth}{6} 
\caption{Summary of Training Set Statistics} 
\label{tab:trainset_stat}

\centering
\resizebox{0.8\linewidth}{!}{ 
\begin{tabular}{c|ccc}
\hline
\bfseries dataset & $N_{train}$ & $M$ & $\frac{N_{train}}{M}$\\ \hline
MNIST & $60000$ & $10$ & $6000$ \\
CIFAR-10 & $50000$ & $10$ & $5000$\\
CIFAR-100 & $50000$ & $100$ &  $500$\\
PubFig83 & $9688$ & $83$ & $\approx 117$ \\
CelebA$_{1000}$ & $25000$ & $1000$ & $25$ \\
\hline
\end{tabular}
}
\end{table}

\para{Results in Transfer Settings}
As in Table \ref{tab:transfer}, while pre-trained features already lift the utility of vanilla DP-SGD, using \name can further improve test accuracy of DP-SGD by $1.23\%\sim 5.81\%$ with $R=R_{\text{best}}$ and $0.56\%\sim 5.81\%$ with $R=10$. 

Firstly, pre-trained features greatly boost capabilities of vanilla DP-SGD in a sense that now it can perform fairly well within meaningful privacy budgets on much more difficult tasks. Table \ref{tab:trainset_stat} contains a summary of training set statistics, where $N_{train}$ denotes the size of training set, $M$ denotes the size of label set and $\frac{N_{train}}{M}$ denotes the average number of samples per label. CIFAR-100, PubFig83 and CelebA$_{1000}$ are considered much harder than MNIST and CIFAR-10, as they have more label classes, fewer samples per class, and richer details in images.
However, with a pre-trained EfficientNet-b0 as the feature extractor, an accuracy of $46.78\%$ is achieved on CIFAR-100 by DP-SGD without smoothing in a very tight privacy budget $\eps=0.86$, which is already comparable to $49.70\%$ on CIFAR-10 by training a CNN from scratch with $\eps=1.99$. Clearly, it is beneficial to have pre-trained features when using DP-SGD.

Secondly, even with strong pre-trained features, randomized smoothing remains a substantial improvement to model performance when applied in DP-SGD. This observation is consistent in all three settings, regardless of whether the feature extractor is pre-trained on a dataset with a broader scope (ImageNet\bm{$\to$}CIFAR-100), a dataset with domain shift (CelebA\bm{$\to$}PubFig83), or even a dataset with a similar distribution (CelebA$_{-}$\bm{$\to$}CelebA$_{1000}$). With the presence of randomized smoothing, test accuracy is raised by at least $3\%$ in many cases and around $5\%$ in a few. The effectiveness of randomized smoothing in improving utility does not seem to be weakened by access to pre-trained features.

\begin{table*}[!t]  
\renewcommand{\arraystretch}{1.3} 
\newcommand{\tablewidth}{6} 
\caption{Evaluation of Next Word Prediction on Reddit} 
\label{tab:reddit}

\centering
\begin{tabular}{|c||c||c||l|l|l|l}
\hline
\multicolumn{1}{|c||}{\multirow{2}{*}{\bfseries dataset}} & \multirow{2}{*}{\bfseries model} & \multirow{2}{*}{\bfseries smoothing} & \multicolumn{ 3 }{c|}{\bfseries perplexity} \\ \cline{4-\tablewidth} 
\multicolumn{1}{|c||}{} & & & \multicolumn{1}{c|}{$\eps=0.55$} & \multicolumn{1}{c|}{$\eps=0.60$} & \multicolumn{1}{c|}{$\eps=0.65$} \\
\hline\hline
\multicolumn{1}{|c||}{\multirow{6}{*}{Reddit}} & \multirow{3}{*}{MLP}& No & $692.85$ & $654.66$ & $652.30$ \\ \cline{3-\tablewidth}
\multicolumn{1}{|c||}{}& & $R=10$ & $638.10\color{blue}{(-54.75)}$ & $593.92\color{blue}{(-60.74)}$ & $590.11\color{blue}{(-62.19)}$  \\ \cline{3-\tablewidth}
\multicolumn{1}{|c||}{}& & $R=R_{\text{best}}$ & $622.81\color{blue}{(-70.04)}$ & $588.76\color{blue}{(-65.90)}$ & $585.50\color{blue}{(-66.80)}$  \\ \cline{2-\tablewidth}
\multicolumn{1}{|c||}{}& \multirow{3}{*}{LSTM} & No & $1157.42$ & $2909.73$ & $2981.27$ \\ \cline{3-\tablewidth}
\multicolumn{1}{|c||}{}& & $R=10$ & $688.95\color{blue}{(-468.47)}$ & $585.47\color{blue}{(-2324.26)}$ & $578.63\color{blue}{(-2402.64)}$ \\ \cline{3-\tablewidth}
\multicolumn{1}{|c||}{}& & $R=R_{\text{best}}$ &  $683.19\color{blue}{(-474.23)}$ & $583.23\color{blue}{(-2326.50)}$ & $575.54\color{blue}{(-2405.73)}$  \\ 
\hline
\end{tabular}
\end{table*}

\para{Results on Next Word Prediction}
In Table~\ref{tab:reddit}, applying \name results in a significant performance improvement, which further supports the effectiveness of our approach. Perplexity is reduced by at least $50$ in all cases for both $R=10$ and $R=R_{\text{best}}$. 

Similar to the image-domain tasks in non-transfer settings, we observe degrading performance for the LSTM model when using DP-SGD without smoothing. The perplexity is indeed increasing gradually when training proceeds. While we will further discuss the potential cause of this issue in Section~\ref{sec:justification_smoothing}, we can see here that it is again addressed by randomized smoothing successfully. Perplexity decreases and utility improves through training with the presence of randomized smoothing.

\subsection{Hyper-parameters for Randomized Smoothing}
Applying randomized smoothing introduces only two scalar hyper-parameters, smoothing radius $R$ and the number of smoothing samples $K$. With existing private hyper-parameter selection techniques\cite{PrivateSelect,PrivateSelect2}, adding two additional scalar hyper-parameters to existing hyper-parameters in DP-SGD is in fact not too much of a burden on privacy budgets, even if they are treated naively through grid search. Nevertheless, we will show in this section that by digging into the effects of $R$ and $K$ to model performance, their selection can be done properly with less or even no additional privacy budget.


\begin{figure}[!t]
	\centering
	\subfloat[MNIST]{
	\includegraphics[width=.45\linewidth]{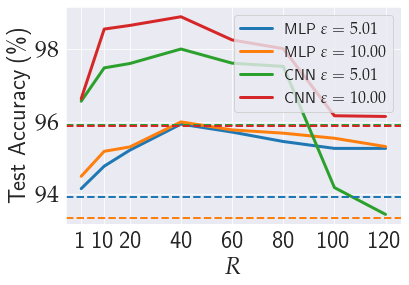}
	\label{fig:radius-mnist}
    }
    \hfil
    \subfloat[CIFAR-10]{
	\includegraphics[width=.45\linewidth]{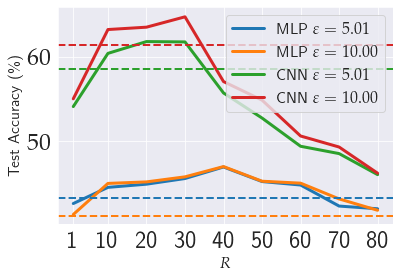}
	\label{fig:radius-cifar}
	}
	
	\subfloat[ImageNet\bm{$\to$}CIFAR-100]{
    \includegraphics[width=0.45\linewidth]{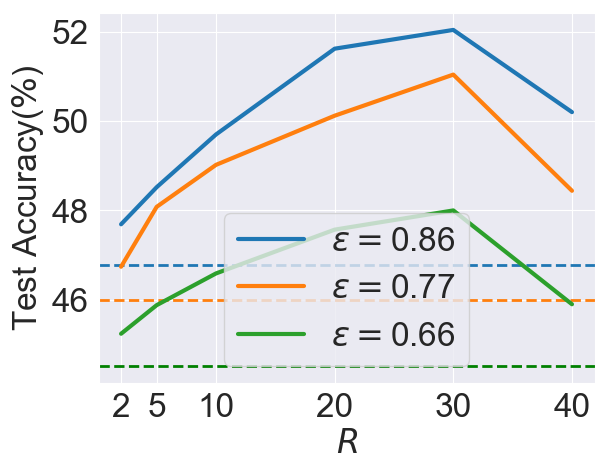} 
    \label{fig:ablation_r_imgnet_cifar100}
    }
    \hfil
    \subfloat[CelebA\bm{$\to$}PubFig83]{
    \includegraphics[width=0.47\linewidth]{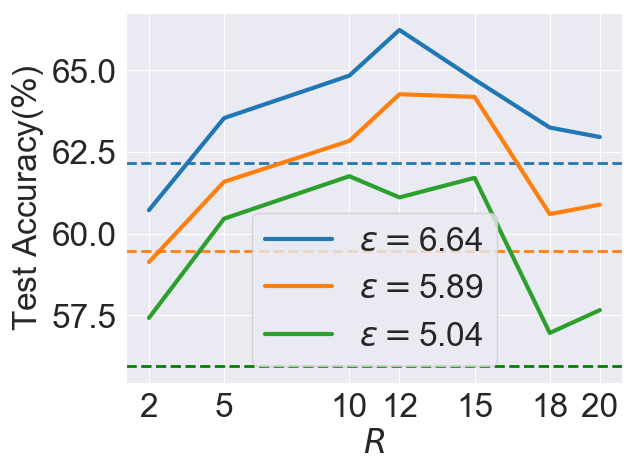}
    \label{fig:ablation_r_celeba_pubfig83}
    }
    
    \subfloat[CelebA$_{-}$\bm{$\to$}CelebA$_{1000}$]{
    \includegraphics[width=0.45\linewidth]{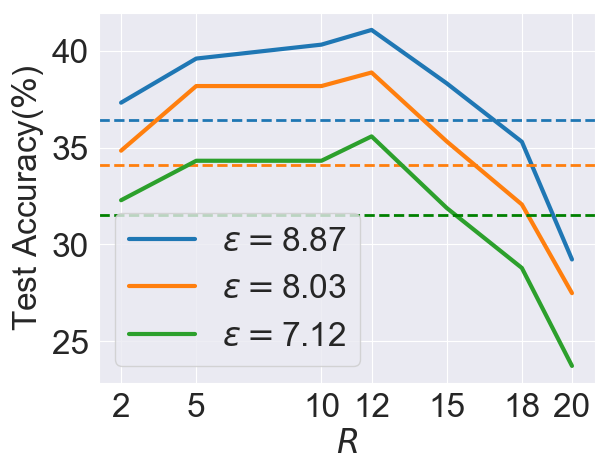}
    \label{fig:ablation_r_celeba_celeba}
    }
    \hfil
    \subfloat[Reddit]{
    \includegraphics[width=0.49\linewidth]{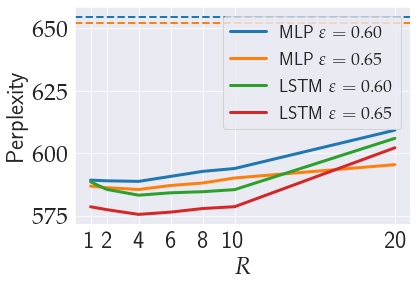}
    \label{fig:radius-reddit}
    }
	\caption{Test accuracy v.s. smoothing radius $R$}
	\label{fig:radius-vs-acc}
\end{figure}

\para{Effect of \bm{$R$}}
\label{sec:effect_radius}
To understand the effect of $R$, we evaluate randomized smoothing with various $R$ under $K=20$ for next word prediction on Reddit and under $K=10$ for other settings.
The results are presented in Figure~\ref{fig:radius-vs-acc}. We also report the values of $R_{\text{best}}$ in Table 
\ref{tab:non-transfer_r_best}, Table \ref{tab:transfer_r_best} and Table \ref{tab:reddit_r_best} respectively for different settings. Each color corresponds to a specific setting with a specific value of $\eps$, with the full line in that color denoting performance with smoothing using different $R$ and the dotted line in that color denoting performance without smoothing.

\begin{table}[!t]  
\renewcommand{\arraystretch}{1.3} 
\newcommand{\tablewidth}{6} 
\caption{$R_{\text{best}}$ in Non-transfer Settings} 
\label{tab:non-transfer_r_best}

\centering
\resizebox{\linewidth}{!}{ 
\begin{tabular}{|c||c||c|c|c|c|l}
\hline
\multicolumn{1}{|c||}{\multirow{2}{*}{\bfseries dataset}} & \multirow{2}{*}{\bfseries model}  & \multicolumn{ 4 }{c|}{\bm{$R_{\text{best}}$}} \\ \cline{3-\tablewidth} 
\multicolumn{1}{|c||}{} & & \multicolumn{1}{c|}{$\eps=1.99$} & \multicolumn{1}{c|}{$\eps=5.01$} & \multicolumn{1}{c|}{$\eps=7.01$} & \multicolumn{1}{c|}{$\eps=10.00$} \\
\hline\hline
\multicolumn{1}{|c||}{\multirow{2}{*}{MNIST}} & \multirow{1}{*}{MLP}&  $40$ & $40$ & $40$ & $40$\\ \cline{2-\tablewidth}
\multicolumn{1}{|c||}{}& \multirow{1}{*}{CNN} & $40$ & $40$ & $40$ & $40$\\ \cline{3-\tablewidth}
\hline\hline
\multicolumn{1}{|c||}{\multirow{2}{*}{CIFAR-10}} & \multirow{1}{*}{MLP} & $30$ & $40$ & $40$ & $40$\\ \cline{2-\tablewidth}
\multicolumn{1}{|c||}{}& \multirow{1}{*}{CNN} & $30$ & $30$ & $30$ & $30$\\ \cline{3-\tablewidth}
\hline
\end{tabular}
}
\end{table}

\begin{table}[!t]  
\renewcommand{\arraystretch}{1.3} 
\newcommand{\tablewidth}{6} 
\caption{$R_{\text{best}}$ in Transfer Settings} 
\label{tab:transfer_r_best}

\centering
\resizebox{\linewidth}{!}{ 
\begin{tabular}{|c||c|c|c|c|c|l}
\hline
\multicolumn{1}{|c||}{\bfseries setting} & \multicolumn{ 5 }{c|}{\bm{$R_{\text{best}}$}} \\ \hline\hline
\multicolumn{1}{|c||}{\multirow{2}{*}{\shortstack{ImageNet\bm{$\to$}\\CIFAR-100}}} & \multicolumn{1}{c|}{$\eps=0.39$} & \multicolumn{1}{c|}{$\eps=0.54$} & \multicolumn{1}{c|}{$\eps=0.66$} & \multicolumn{1}{c|}{$\eps=0.77$} & \multicolumn{1}{c|}{$\eps=0.86$} \\ \cline{2-\tablewidth}
\multicolumn{1}{|c||}{}& $5$ & $12$ & $30$ & $30$ & $30$\\
\hline
\hline
\multicolumn{1}{|c||}{\multirow{2}{*}{\shortstack{CelebA\bm{$\to$}\\PubFig83}}} & \multicolumn{1}{c|}{$\eps=2.84$} & \multicolumn{1}{c|}{$\eps=4.06$} & \multicolumn{1}{c|}{$\eps=5.04$} & \multicolumn{1}{c|}{$\eps=5.89$} & \multicolumn{1}{c|}{$\eps=6.64$} \\ \cline{2-\tablewidth}
\multicolumn{1}{|c||}{}& $15$ & $12$ & $10$ & $12$ & $12$\\
\hline\hline
\multicolumn{1}{|c||}{\multirow{2}{*}{\shortstack{CelebA$_{-}$\bm{$\to$}\\CelebA$_{1000}$}}} & \multicolumn{1}{c|}{$\eps=4.88$} & \multicolumn{1}{c|}{$\eps=6.06$} & \multicolumn{1}{c|}{$\eps=7.12$} & \multicolumn{1}{c|}{$\eps=8.03$} & \multicolumn{1}{c|}{$\eps=8.87$} \\ \cline{2-\tablewidth}
\multicolumn{1}{|c||}{}& $12$ & $12$ & $12$ & $12$ & $12$\\ 
\hline
\end{tabular}
}
\end{table}

\begin{table}[!t]  
\renewcommand{\arraystretch}{1.3} 
\newcommand{\tablewidth}{5} 
\caption{$R_{\text{best}}$ for Next Word Prediction on Reddit} 
\label{tab:reddit_r_best}

\centering
\begin{tabular}{|c||c||c|c|c|l}
\hline
\multicolumn{1}{|c||}{\multirow{2}{*}{\bfseries dataset}} & \multirow{2}{*}{\bfseries model}  & \multicolumn{ 3 }{c|}{\bm{$R_{\text{best}}$}} \\ \cline{3-\tablewidth} 
\multicolumn{1}{|c||}{} & & \multicolumn{1}{c|}{$\eps=0.55$} & \multicolumn{1}{c|}{$\eps=0.60$} & \multicolumn{1}{c|}{$\eps=0.65$} \\
\hline\hline
\multicolumn{1}{|c||}{\multirow{2}{*}{Reddit}} & \multirow{1}{*}{MLP}& $4$ & $4$ & $4$ \\ \cline{2-\tablewidth}
\multicolumn{1}{|c||}{}& \multirow{1}{*}{LSTM} & $4$ & $4$ & $4$ \\ \cline{3-\tablewidth}
\hline
\end{tabular}
\end{table}

Regarding the effect of $R$ to model performance, we observe a useful property here: model performance is approximately a unimodal function of $R$, which means it is monotonically increasing for $R<R_{\text{best}}$ and monotonically decreasing for $R>R_{\text{best}}$. With this property, the set of $R$ that yields considerable improvements forms an interval that is always fairly wide in all our results, which is friendly to private hyper-parameter selection. 

Furthermore, benefiting from our factorization of $\sigma_{\text{smooth}}$, even though $\sigma_{\text{smooth}}$ may vary greatly, the interval for $R$ with solid improvements is more or less of the same scale across different settings. As a result, while $R_{\text{best}}$ is different from one setting to another, smoothing DP-SGD with a single radius setting $R=10$ suffices to considerably outperform vanilla DP-SGD across all settings as shown in Table \ref{tab:non-transfer}, Table \ref{tab:transfer} and Table \ref{tab:reddit}. Thus, as a rule of thumb, simply selecting $R=10$ is likely a good choice.

\begin{figure}[!t]
\centering 
\subfloat[ImageNet\bm{$\to$}CIFAR-100]{
\includegraphics[width=0.45\linewidth]{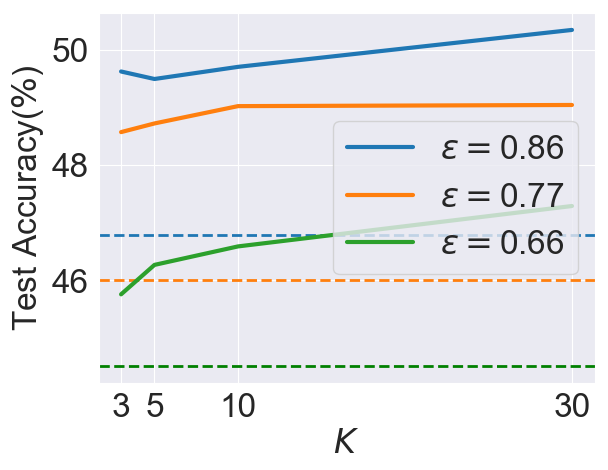} 
\label{fig:ablation_K_imgnet_cifar100}
}
\hfil
\subfloat[CelebA\bm{$\to$}PubFig83]{
\includegraphics[width=0.45\linewidth]{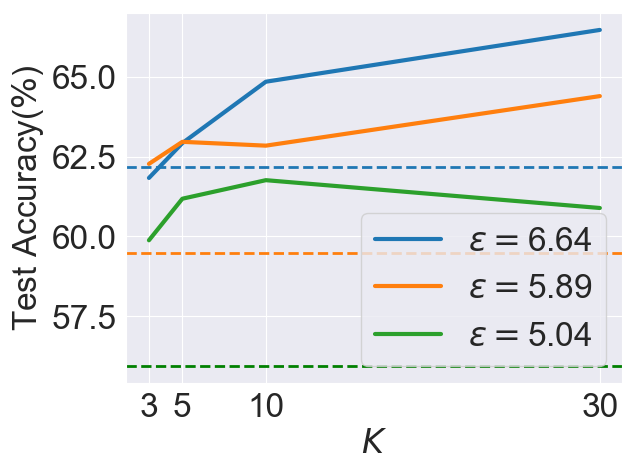}
\label{fig:ablation_K_celeba_pubfig83}
}
\caption{Test accuracy v.s. number of smoothing samples K}
\label{fig:ablation_K} 
\end{figure}

\begin{figure}[!t]
\centering 
\subfloat[no smoothing]{
\includegraphics[width=0.45\linewidth]{images/threeDsurface/nonsmooth_101.png} 
\label{fig:3dsurface-nonsmooth}
}
\hfil
\subfloat[$K=5$]{
\includegraphics[width=0.45\linewidth]{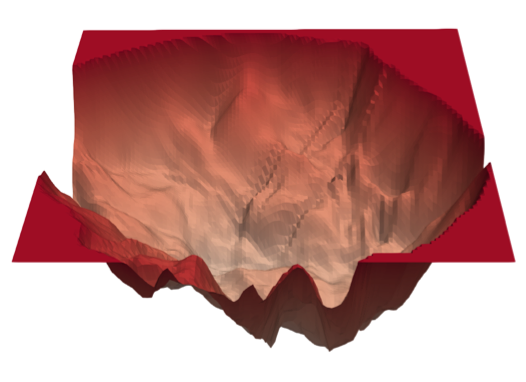} 
\label{fig:3dsurface-k5}
}

\subfloat[$K=10$]{
\includegraphics[width=0.45\linewidth]{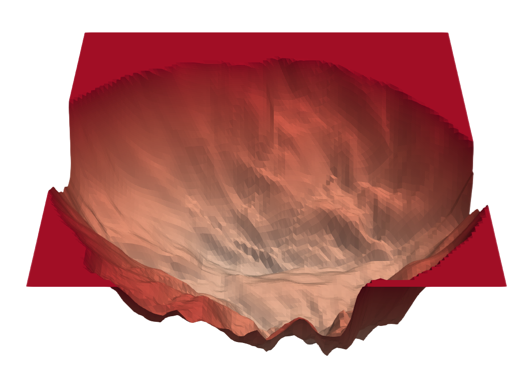} 
\label{fig:3dsurface-k10}
}
\hfil
\subfloat[$K=20$]{
\includegraphics[width=0.45\linewidth]{images/threeDsurface/K20_101.png} 
\label{fig:3dsurface-k20}
}
\caption{Loss surfaces on $1\%$ training data of CIFAR-10}
\label{fig:3dsurface} 
\end{figure}

\para{Effect of \bm{$K$}}
To understand the role of the hyper-parameter $K$, we evaluate randomized smoothing with various $K$ under a fixed radius $R=10$ in two settings, ImageNet\bm{$\to$}CIFAR-100 for object classification and CelebA\bm{$\to$}PubFig83 for face identification. The results are presented in Figure \ref{fig:ablation_K}. 
For each setting, we present results corresponding to three different privacy budget $\eps$. 
For each $\eps$, the test accuracy obtained with different $K$ is plotted as a full line and the test accuracy without smoothing under the same $\eps$ is plotted as a horizontal dotted line in the same color. 

Overall, model performance benefits from a larger $K$, which makes the selection of $K$ a simple trade-off between utility and computation time. Such trade-off can be easily addressed without tuning by selecting the largest $K$ with acceptable training time. This phenomenon is fairly intuitive given that a larger $K$ implies a better approximation of $\calL_\text{smooth}(\theta)$. Figure \ref{fig:3dsurface} supports this claim, in which we visualize the effect of $K$ to the loss surface of a ResNet-56 published by \cite{VisualLandscape} along with 3D loss visualization code\myendnote{https://github.com/tomgoldstein/loss-landscape}{note:landscape}. Besides, even with $K$ as small as 3, randomized smoothing can already improve the performance of DP-SGD considerably, as shown in Figure \ref{fig:ablation_K}.

\subsection{Justification of Smoothing}
\label{sec:justification_smoothing}

\begin{figure}[!t]
\centering 
\subfloat[MNIST MLP]{
\includegraphics[width=0.48\linewidth]{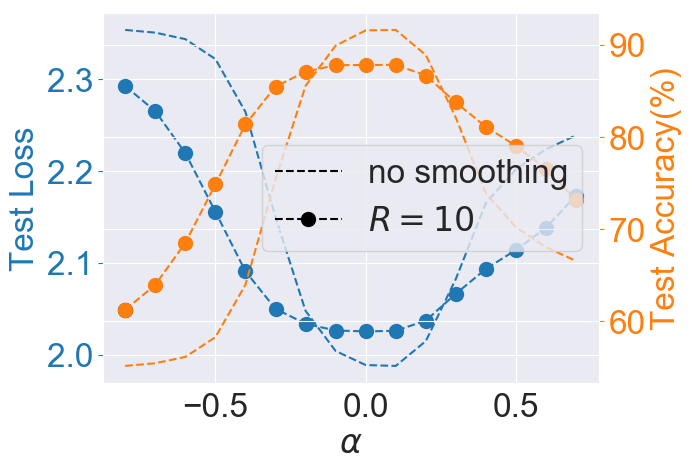} 
\label{fig:sharp-mnist-mlp}
}
\subfloat[MNIST CNN]{
\includegraphics[width=0.48\linewidth]{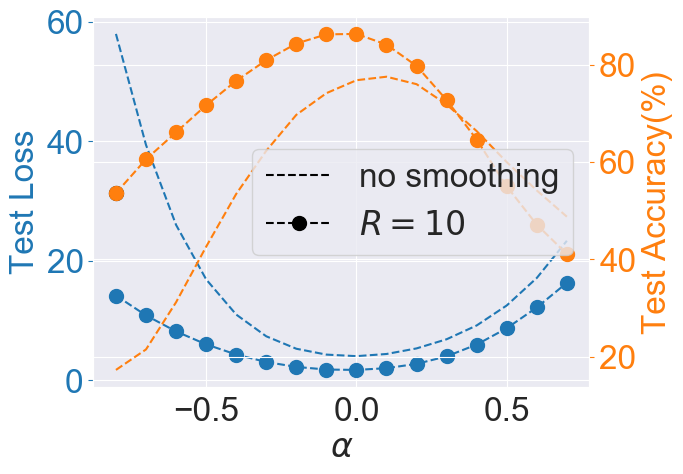}
\label{fig:sharp-mnist-cnn}
}

\subfloat[CIFAR-10 MLP]{
\includegraphics[width=0.48\linewidth]{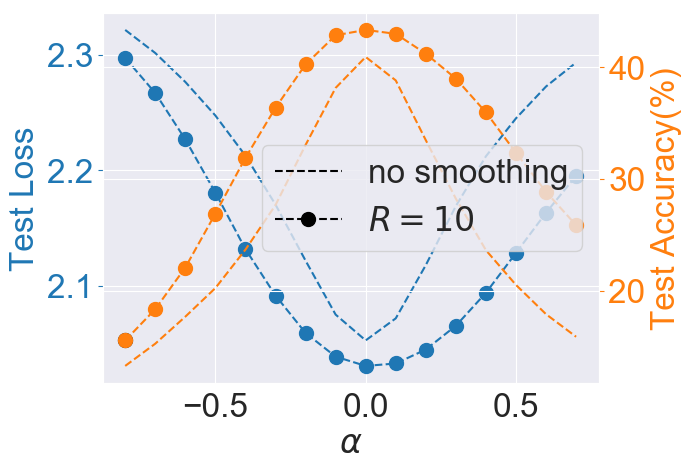}

\label{fig:sharp-cifar-mlp}
}
\subfloat[CIFAR-10 CNN]{
\includegraphics[width=0.48\linewidth]{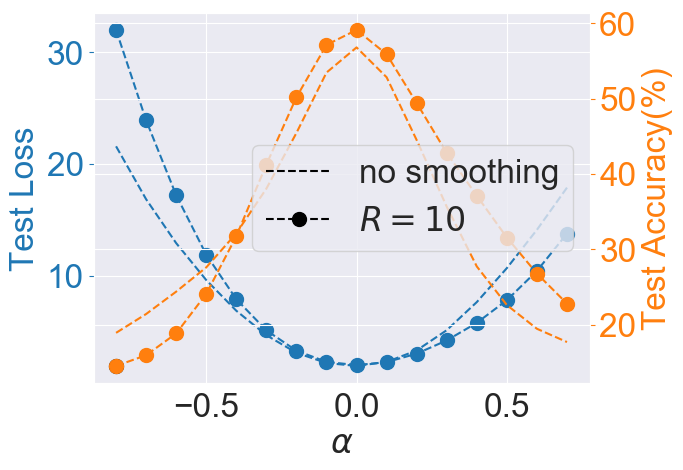}
\label{fig:sharp-cifar-cnn}
}
\caption{Sharpness visualization}
\label{fig:visual_sharpness} 
\end{figure}

When designing \name, one of our motivations is to explicitly incorporate the preference to flatter regions, which is supported by theoretical results in Section \ref{sec:theory}. In this section, we provide further supports by examining empirically whether DP-SGD is guided towards flatter regions with randomized smoothing.

One way of examination is through visualization, which is intuitive but considers limited dimensions. We adapt filter normalized plots from \cite{keskar2016large} for visualization, which is designed to remove apparent differences in geometry caused by scaling filter weights. To visualize the sharpness of loss landscape around a given parameter $\theta$, one begins with a random Gaussian direction $d$ with the same dimensions as $\theta$. Then each filter $f_d$ in $d$ will be normalized by a factor of $\frac{\| f_d \|}{\| f_\theta \|}$, so that it will have the same norm as the corresponding filter $f_\theta$ in $\theta$. At last, with the filter-normalized vector $d$, one can plot out test accuracy and test loss corresponding to models parameters on the segment $\theta + \alpha \cdot d$ with $\alpha \in [-0.8, 0.8]$.

We visualize models trained with and without randomized smoothing in three different settings. The results are presented in Figure \ref{fig:visual_sharpness}. Overall, the visualized region around the parameter obtained by DP-SGD with smoothing tends to be flatter than the one around the parameter obtained by vanilla DP-SGD. Though the visualization is conducted along a single direction, it gives us a rough idea of the effect of \name.

Another way of examination is through numerical sharpness metrics, which is less intuitive but able to capture patterns from more dimensions. Here, we follow the definition of $(C_{\epsilon}; A)$-sharpness from \cite{keskar2016large}.
\begin{myDef}
\textbf{\bm{$(C_{\epsilon}; A)$}-sharpness}: Given $\theta \in \mathcal{R}^{n}$, $\epsilon>0$ and $A\in \mathcal{R}^{n\times p}$, the $(C_{\epsilon}; A)$-sharpness of $\calL$ at $\theta$ is
\begin{equation*}
    \frac{\left( \max_{y\in C_{\epsilon}(\theta)} \calL(\theta + Ay) \right) - \calL(\theta)}{1 + \calL(\theta)} \times 100,
\end{equation*}
where 
$
    C_{\epsilon}(\theta) = \{z\in \mathcal{R}^p: 
    |z_i| \leq \epsilon \left( | \left(A^+ \theta\right)_i | + 1 \right)
    \}
$
and $A^+$ denotes pseudo-inverse of $A$.
\end{myDef}

We set $A$ to be the identity matrix $I_n$ following their default setting and set $\epsilon$ to be $0.0045$. The results are presented in Table \ref{tab:sharp-compu}, where $(C_{\epsilon}; A)$-sharpness is always smaller with randomized smoothing, suggesting the success of our design in guiding DP-SGD towards favorable flatter regions.

\begin{table}[]
	\caption{$(C_{\epsilon}; A)$-sharpness}
	\centering
	\resizebox{\linewidth}{!}{
	\begin{tabular}{|c||c||c||c|l}
		\hline
		\multicolumn{1}{|c||}{\bfseries dataset}   & \bfseries  model & \bfseries  smoothing & \bfseries  \bm{$(C_{\epsilon}; A)$}-sharpness\\ \hline \hline
		\multicolumn{1}{|c||}{\multirow{4}{*}{MNIST}}      & \multirow{2}{*}{MLP}   & No & $0.8626$  \\ \cline{3-4}
		\multicolumn{1}{|c||}{} & & $R=10$ & 0.5454 \\ \cline{2-4}
		\multicolumn{1}{|c||}{}  & \multirow{2}{*}{CNN}  & No   & $139.8159$ \\\cline{3-4}
		\multicolumn{1}{|c||}{}&& $R=10$ & $49.6169$  \\ \hline
		\multicolumn{1}{|c||}{\multirow{4}{*}{CIFAR-10}} & \multirow{2}{*}{MLP} & No & $1.0375$\\\cline{3-4}
		\multicolumn{1}{|c||}{}& & $R=10$ & $0.0000$ \\ \cline{2-4}
		 \multicolumn{1}{|c||}{} & \multirow{2}{*}{CNN}  & No   & $563.5351$ \\\cline{3-4}
		\multicolumn{1}{|c||}{}&& $R=10$ & $340.5471$  \\ 
		\hline
	\end{tabular}
	}
    \label{tab:sharp-compu}
\end{table}

\begin{figure}[!t]
	\centering
	\subfloat[Training accuracy v.s. $\eps$]{
		\includegraphics[width=.4\linewidth]{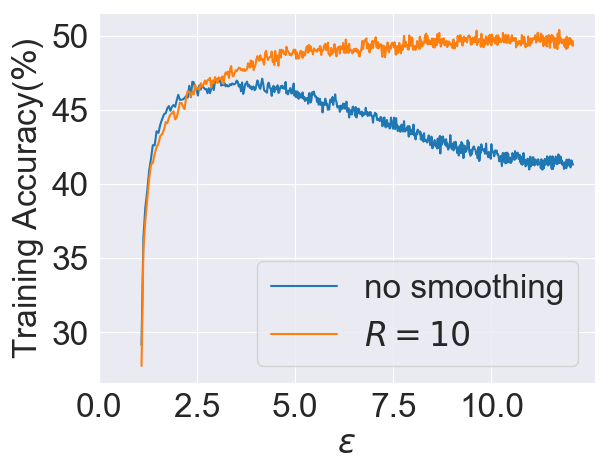}
	\label{fig:degrading_train_acc}
	}
	\hfil
    \subfloat[Sharpness visualization of SGD]{
    	\includegraphics[width=.45\linewidth]{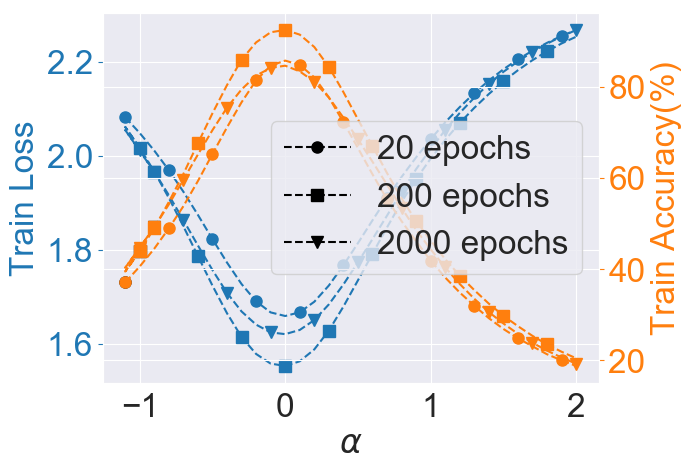}
    \label{fig:SGD-sharpening}
    }
    \label{fig:degrading_test_acc}
    \caption{An explanation to the cause of degrading performance issue of DP-SGD}
\end{figure}

In Section \ref{sec:performance_eval}, we briefly mention the degrading performance issue of vanilla DP-SGD when it is observed in Table \ref{tab:non-transfer} and Table \ref{tab:reddit}, where model performance decreases gradually as training proceeds. Here we propose an explanation for the main cause of such an issue, which separates it from overfitting in machine learning, and we argue that guiding DP-SGD towards flatter regions via randomized smoothing can indeed resolve this issue.

In Figure \ref{fig:degrading_train_acc}, we have curves of training accuracy on CIFAR-10 plotted for DP-SGD with and without randomized smoothing. We notice that not only the performance on the test set but also the performance on the training set can be degrading when using vanilla DP-SGD. The degrading training performance, instead of the enlarged generalization gap (i.e., overfitting), is the major cause for degrading test performance. 

We further propose a hypothesis: degrading training performance is a joint effect of a growing favor to sharper minima and the noisy nature of DP-SGD. In Figure \ref{fig:SGD-sharpening}, we visualize training accuracy and training loss of three checkpoints from a training process with SGD and gradient clipping, where we observe that SGD may start guiding the parameter towards sharper minima from some point of training. Sharper minima are considered less noise-tolerant and degrading training accuracy occurs when starting at some point they fail to tolerate noise introduced by DP-SGD. Randomized smoothing resolves this issue by mitigating the tendency for sharper minima with an explicit preference to flatter regions.












\begin{table*}[!t]
\caption{Stability of Performance}
\label{tb:stability}
\centering
\resizebox{0.9\linewidth}{!}{ 
\renewcommand{\arraystretch}{1.3} 
\begin{tabular}{|c||c||c||c|c||c|c|l}
\hline
\multicolumn{1}{|c||}{\multirow{2}{*}{\textbf{dataset}}} & \multirow{2}{*}{\textbf{model}} & \multirow{2}{*}{\textbf{smoothing}} & \multicolumn{2}{c||}{\textbf{standard   deviation}} & \multicolumn{2}{c|}{\textbf{range}}                       \\ \cline{4-7} 
\multicolumn{1}{|c||}{}                                  &                                 &                                     & $\eps=5.01$                & $\eps=10.00$                & $\eps=5.01$                    & $\eps=10.00$                   \\ \hline \hline
\multicolumn{1}{|c||}{\multirow{3}{*}{MNIST}}            & \multirow{3}{*}{MLP}            & No                                  & 0.184\%                 & 0.339\%                  & 0.55\% (93.51\%,   94.06\%) & 1.02\% (93.25\%,   94.27\%) \\ \cline{3-7} 
\multicolumn{1}{|c||}{}                                  &                                 & $R=10$                              & 0.088\%                 & 0.184\%                  & 0.25\% (94.72\%,   94.97\%) & 0.56\% (95.36\%,   95.92\%) \\ \cline{3-7} 
\multicolumn{1}{|c||}{}                                  &                                 & $R=40$                              & 0.068\%                 & 0.088\%                  & 0.18\% (95.92\%,   96.10\%) & 0.25\% (95.90\%,   96.15\%) \\ 
\hline \hline
\multicolumn{1}{|c||}{\multirow{3}{*}{CIFAR-10}}          & \multirow{3}{*}{CNN}            & No                                  & 0.577\%                 & 0.427\%                  & 1.61\% (57.27\%,   58.88\%) & 1.18\% (60.65\%,   61.83\%) \\ \cline{3-7} 
\multicolumn{1}{|c||}{}                                  &                                 & $R=10$                              & 0.444\%                 & 0.360\%                  & 1.24\% (60.18\%,   61.42\%) & 0.92\% (62.66\%, 63.58\%)   \\ \cline{3-7} 
\multicolumn{1}{|c||}{}                                  &                                 & $R=30$                              & 0.332\%                 & 0.196\%                  & 0.96\% (61.22\%, 62.18\%)   & 0.58\% (64.25\%,   64.83\%) \\ \hline
\end{tabular}
}
\end{table*}

\begin{table*}[!t]  
\renewcommand{\arraystretch}{1.3} 
\newcommand{\tablewidth}{7} 
\caption{Evaluation when combined with Tempered Sigmoid} 
\label{tab:tempered_sigmoid}

\centering
\resizebox{0.9\linewidth}{!}{ 
\begin{tabular}{|c||c||c||l|l|l|l|l}
\hline
\multicolumn{1}{|c||}{\multirow{2}{*}{\bfseries dataset}} & \multirow{2}{*}{\bfseries model} & \multirow{2}{*}{\bfseries smoothing} & \multicolumn{ 4 }{c|}{\bfseries test accuracy} \\ \cline{4-\tablewidth} 
\multicolumn{1}{|c||}{} & & & \multicolumn{1}{c|}{$\eps=1.99$} & \multicolumn{1}{c|}{$\eps=5.01$} & \multicolumn{1}{c|}{$\eps=7.01$} & \multicolumn{1}{c|}{$\eps=10.00$} \\
\hline\hline
\multicolumn{1}{|c||}{\multirow{3}{*}{CIFAR-10}}& \multirow{3}{*}{CNN} & No & $50.04\%$ & $57.01\%$ & $58.07\%$ & $57.46\%$\\ \cline{3-\tablewidth}
\multicolumn{1}{|c||}{}& & $R=10$ & $50.38\%\color{blue}{(+0.34\%)}$ & $58.18\%\color{blue}{(+1.17\%)}$ & $59.58\%\color{blue}{(+1.51\%)}$ & $59.13\%\color{blue}{(+1.67\%)}$ \\ \cline{3-\tablewidth}
\multicolumn{1}{|c||}{}& & $R=R_{\text{best}}$ & $50.38\%\color{blue}{(+0.34\%)}$ & $59.17\%\color{blue}{(+2.16\%)}$ & $59.93\%\color{blue}{(+1.86\%)}$ & $60.06\%\color{blue}{(+2.60\%)}$ \\ 
\hline
\end{tabular}
}
\end{table*}

\begin{table*}[!t]  
\renewcommand{\arraystretch}{1.3} 
\newcommand{\tablewidth}{7} 
\caption{Evaluation of Smoothing for PATE} 
\label{tab:PATE}

\centering
\resizebox{0.9\linewidth}{!}{ 
\begin{tabular}{|c||c||c||l|l|l|l|l}
\hline \multicolumn{1}{|c||}{\bfseries dataset} &
\bfseries parameter & \bfseries smoothing & \multicolumn{ 4 }{c|}{\bfseries test accuracy} \\ \hline\hline
\multicolumn{1}{|c||}{\multirow{9}{*}{SVHN}} & \multirow{3}{*}{$\sigma_2=40$} &  & \multicolumn{1}{c|}{$\eps=7.44$} & \multicolumn{1}{c|}{$\eps=11.08$} & \multicolumn{1}{c|}{$\eps=14.08$} & \multicolumn{1}{c|}{$\eps=16.75$} \\ \cline{3-\tablewidth}
\multicolumn{1}{|c||}{}& &  No &  $82.56\%$ & $84.44\%$ & $85.79\%$ & $86.77\%$\\ \cline{3-\tablewidth}
\multicolumn{1}{|c||}{}& & Yes & $84.06\%\color{blue}{(+1.50\%)}$ & $85.64\%\color{blue}{(+1.20\%)}$ & $86.96\%\color{blue}{(+1.17\%)}$ & $87.39\%\color{blue}{(+0.62\%)}$ \\
\cline{2-\tablewidth}
\multicolumn{1}{|c||}{}& \multirow{3}{*}{$\sigma_2=80$} &  & \multicolumn{1}{c|}{$\eps=3.78$} & \multicolumn{1}{c|}{$\eps=5.51$} & \multicolumn{1}{c|}{$\eps=6.90$} & \multicolumn{1}{c|}{$\eps=8.10$} \\ \cline{3-\tablewidth}
\multicolumn{1}{|c||}{}& &  No &  $75.65\%$ & $76.77\%$ & $81.95\%$ & $83.28\%$\\ \cline{3-\tablewidth}
\multicolumn{1}{|c||}{}& & Yes & $77.84\%\color{blue}{(+2.19\%)}$ & $82.06\%\color{blue}{(+5.29\%)}$ & $83.62\%\color{blue}{(+1.67\%)}$ & $84.64\%\color{blue}{(+1.36\%)}$ \\
\cline{2-\tablewidth}
\multicolumn{1}{|c||}{}& \multirow{3}{*}{$\sigma_2=100$} &  & \multicolumn{1}{c|}{$\eps=3.15$} & \multicolumn{1}{c|}{$\eps=4.57$} & \multicolumn{1}{c|}{$\eps=5.70$} & \multicolumn{1}{c|}{$\eps=6.68$} \\ \cline{3-\tablewidth}
\multicolumn{1}{|c||}{}& &  No &  $70.51\%$ & $75.75\%$ & $79.19\%$ & $80.50\%$\\ \cline{3-\tablewidth}
\multicolumn{1}{|c||}{}& & Yes & $73.92\%\color{blue}{(+3.41\%)}$ & $79.56\%\color{blue}{(+3.81\%)}$ & $81.69\%\color{blue}{(+2.50\%)}$ & $82.36\%\color{blue}{(+1.86\%)}$ \\
\hline
\end{tabular}
}
\end{table*}

\subsection{Stability of Performance}

Vanilla DP-SGD is usually considered unstable as variations of performance among independent runs are non-negligible. In this section, we show that randomized smoothing improves not only the performance of DP-SGD, but also its stability.

Table \ref{tb:stability} contains our evaluation results. For each cell in the table, range and standard deviation of test accuracy across 5 runs are reported. Both standard deviation and range can be reduced significantly after using randomized smoothing, which indicates that randomized smoothing helps with the stability of performance.

\subsection{Compatibility of \name}
Since \name tailors the training objectives through smoothing, it is by design complementary to and therefore naturally compatible with most if not all existing strategies. As a proof of concept, we show empirically in this section that the gains from \name remain when combined with Tempered Sigmoid \cite{TemperedSigmoid}, which modifies the activation functions.

The results are included in Table \ref{tab:tempered_sigmoid}. Following \cite{TemperedSigmoid}, we replace ReLU activations with tanh, the default setting of Tempered Sigmoid. Other setups are the same as in Section \ref{sec:evaluationsetup}.
The values of $R_\text{best}$ are $10, 15, 20, 15$ respectively for $\eps=1.99, 5.01, 7.01, 10.00$.

On CIFAR-10 using a CNN with Tempered Sigmoid activations, \name improves the test accuracy by $0.34\%\sim 2.60\%$ with $R=R_\text{best}$ and $0.34\%\sim 1.67\%$ with $R=10$, which corroborates the compatibility of \name with other strategies.

\subsection{Extension to PATE}

In this section, we show that the idea of smoothing learning loss can be extended easily to PATE for improved utility as well. 
The core of PATE is noisy data labeling. Intuitively, the label noise manifests itself as parameter noise during training, which can be mitigated by smoothing.

\para{Experimental Setup}
We evaluate PATE on SVHN\cite{SVHN} benchmark following mostly setups from \cite{papernot2016semi, papernot2018scalable}. Detailed experimental setup is included in Appendix \ref{sec:setup_PATE}.

\para{Experimental Results}
As shown in Table \ref{tab:PATE}, smoothing improves test accuracy of PATE by $0.62\%\sim 5.29\%$. PATE benefits from smoothing because of the increased tolerance to the noise in privacy-preserving labels. In Table \ref{tab:PATE_details}, we report the accuracy of privacy-preserving labeling $P_{\text{correct}}$. We notice that in most cases, the improvement from smoothing is greater for settings with lower $P_{\text{correct}}$, which corroborates the ability of smoothing to handle label noise.

\section{Related Work}
\label{sec:related}





\para{Privacy Attacks}
Various privacy attacks motivate the development of DP. 
Privacy attacks against machine learning models aim to learn private information about the training data. Membership inference and model inversion are two major attacks of interest in the literature. Membership inference attacks aim to determine whether a given individual’s data record is included in the training set of the model \cite{shokri2017membership, salem2018ml, long2018understanding, li2020label, chen2019gan}. On the other hand, model inversion attacks aim to reconstruct the sensitive features of the training data records \cite{fredrikson2014privacy, fredrikson2015model, yang2019adversarial, salem2020updates, zhang2020secret}. Additionally, model inversion attack can be further refined into property inference attack, where the attacker can speculate whether there is a certain statistical property in the training dataset \cite{ateniese2015hacking, ganju2018property, wang2019robust}. While DP provides a strong theoretical guarantee against these privacy attacks, it will typically cause unbearable utility degradation for the trained models \cite{rahman2018membership, wang2020improving}. Our work can be applied to obtain a better utility-privacy tradeoff against privacy attacks.

\para{Differentially private optimization}
Differentially private optimization is one of the most important applications of DP and has gone through careful studies during the past decade.
For convex optimization, there is a line of works with different trade-offs between utility, privacy, and usability.
\cite{chaudhuri2011differentially} proposed the classic techniques
of output perturbation and objective perturbation. 
A thorough analysis of the techniques appear in~\cite{jain2014near} and two variants were proposed in~\cite{wu2017bolt,iyengar2019towards}.
Several mechanisms~\cite{kifer2012private,thakurta2013differentially,talwar2014private} were proposed for dealing with high-dimensional sparse regression.
The non-convex setting has only seen real progress recently.
The first private SGD algorithm was given in~\cite{song2013stochastic}, but only until the emergence of DP-SGD~\cite{DBLP:conf/ccs/AbadiCGMMT016} with improved privacy composition do we see the real deployment of DP in deep learning systems. 
Since then, \cite{bun2016concentrated,mironov2017renyi} proposed Concentrated-DP and R\'enyi-DP which provides better composability for DP-SGD.
\cite{wang2019subsampled} further improved the composition by considering sub-sampling in R\'enyi differential privacy.
\cite{lee2020differentially} proposed to use direct feedback alignment instead of backward propagation in DP-SGD.
\cite{TemperedSigmoid} suggested that a family of bounded activation functions, namely the tempered sigmoids, is more preferable than ReLU in DP-SGD.
\cite{LearnToProtect} used recurrent neural networks learned on auxiliary public data to adjust noise scales and to decide the update direction from the noisy gradient.
\cite{CDP_adaptive_budget} introduced a line-search module to privately choose among a pre-defined set of learning rates and decide whether to allocate more privacy budget. 
In addition, \cite{CDP_adaptive_budget} showed that setting a smaller gradient clipping threshold $C$ may impair utility when it causes too much information loss in the estimates. Besides, unlike randomized smoothing used in this work, reducing $C$ does not affect the smoothness of loss landscapes and hence does not help tolerate noise.

\para{Randomized Smoothing}
Typically, a loss function can be viewed as a function of both model parameters and input data. 
In the machine learning community, several lines of works perform randomized smoothing over model parameter space as we do.
To improve generalization, \cite{chaudhari2019entropy} proposed Entropy-SGD, which optimizes local entropy instead of the original loss function by estimating gradients with Langevin dynamics\cite{LangevinDynamics} and local entropy is designed to have a smooth energy landscape.
With a similar purpose, \cite{wen2018smoothout} proposed SmoothOut, which injects noise into the model to smooth out sharp minima to obtain more robust models with flatter minima.
\cite{duchi2012randomized} utilized parameter-space smoothing to improve convergence rates in nonsmooth convex optimizations.

\section{Conclusion}
\label{sec:conclusion}
In this work, we propose \name, which improves the utility of DP-SGD by smoothing the learning loss function. 
We show both theoretically and empirically that DP-SGD with the smoothed loss not only reduces the performance variation across different runs but also achieves a better generalization bound. 
In addition, we show that the idea of smoothing learning loss can be easily extended to improve the utility of PATE.

Through our work, we hope to open up a new perspective for improving the utility of privacy-preserving deep learning, which is to tailor learning objective functions to differential privacy goals. For future work, we will investigate different instantiations of this overarching idea and study the applications of the proposed approach to medical data analysis.

\section{Acknowledgement}

This work is supported by National Natural Science Foundation of China (NSFC) under grant no. 61972448, DARPA contract \#N66001-15-C-4066, the Center for Long-Term Cybersecurity, and Berkeley Deep Drive. 
\begin{appendices}
\runningtitle{\name: Boosting Utility of Differentially Private Deep Learning via Randomized Smoothing}

\section{Proof of Lemma \ref{lem:smooth}}
\begin{proof}
Let $\mu(\Delta)$ to be the density of a random variable $\Delta\sim \mathcal{N}(0,\sigma_\text{smooth}^2\Id)$, then we have
\begin{align*}
     \calL_\text{smooth}(\theta)& =\mathbb{E}_{\Delta \sim \mathcal{N}(0,\sigma_\text{smooth}^2\Id)}[\calL(\theta+\Delta)]\\
     &= \mathbb{E}_{\theta' \sim \mathcal{N}(\theta,\sigma_\text{smooth}^2\Id)}[\calL(\theta')] \\
     &=\int_{\theta'} \calL(\theta') \mu(\theta - \theta') d\theta'.
\end{align*}
With Lemma 9(iii) from \cite{duchi2012randomized}, given that $\calL$ is $L$-Lipschitz with respect to the $\ell_2$ norm, we have the gradient of $\calL_\text{smooth}$ to be $L/\sigma_\text{smooth}$-Lipschitz continuous and therefore $\calL_\text{smooth}$ is $L/\sigma_\text{smooth}$-smooth.

With Lemma 9(iv) from \cite{duchi2012randomized}, we have 
\begin{align*}
    \nabla \calL_\text{smooth}(\theta) =& \mathbb{E}_{\Delta \sim \mathcal{N}(0,\sigma_\text{smooth}^2\Id)}[\nabla \calL(\theta+\Delta)]\\
    =&\int_{\Delta} \left( \nabla\calL(\theta + \Delta) \right) \mu(\Delta) d\Delta
\end{align*}
Thus when $\calL$ is $\beta$-smooth, for any $\theta, \theta'$, we have
\begin{align*}
    &\| \nabla \calL_\text{smooth}(\theta) - \nabla\calL_\text{smooth}(\theta') \| \\
    =&\| \int_{\Delta} \left( \nabla\calL(\theta + \Delta) \right) \mu(\Delta) d\Delta - \int_{\Delta}  \left(\nabla \calL(\theta' + \Delta)\right) \mu(\Delta) d\Delta \| \\
    =& \|\int_{\Delta} \left( \nabla\calL(\theta + \Delta)  - \nabla \calL(\theta' + \Delta) \right) \mu(\Delta) d\Delta \|\\
    \leq & \max_{\Delta} \| \nabla\calL(\theta + \Delta)  - \nabla \calL(\theta' + \Delta) \|\\
    \leq & \beta \| \theta - \theta' \|,
\end{align*}
which means $\calL_\text{smooth}$ is at least  $\beta$-smooth. 

\end{proof}

\section{Proof of Theorem \ref{cor:conv}}
\begin{proof}
Since $\mathcal{L}$ is $\beta$-smooth,
\begin{equation*}
\begin{split}
\mathcal{L}&(\theta_{t+1}) \leq \mathcal{L}(\theta_t) + \langle\nabla\mathcal{L}(\theta_t), \theta_{t+1}-\theta_{t}\rangle + \frac{\beta}{2}\eta^2\|\nabla\mathcal{L}(\theta_t, x_t)\|^2\\
& = \mathcal{L}(\theta_t) - \eta\langle\nabla\mathcal{L}(\theta_t), \nabla\mathcal{L}(\theta_t, x_t)\rangle + \frac{\beta}{2}\eta^2\|\nabla\mathcal{L}(\theta_t, x_t)\|^2\\
\end{split}
\end{equation*}
Take expectation on both sides, we have
\begin{equation*}
\begin{split}
&\mathbb{E}[\mathcal{L}(\theta_{t+1})] - \mathbb{E}[\mathcal{L}(\theta_t)] \\
&\leq - \eta\|\nabla\mathcal{L}(\theta_t)\|^2 + \frac{\beta}{2}\eta^2\|\nabla\mathcal{L}(\theta_t)\|^2+\frac{\beta}{2}\eta^2\sigma^2\\
\end{split}
\end{equation*}
By summarizing the above equation through all time steps, we obtain the following inequality.
\begin{equation*}
\begin{split}
    \frac{1}{T}\sum_{t=1}^T\mathbb{E}[\|\nabla \calL(\theta_t)\|^2] &\leq  \frac{2\mathbb{E}[\mathcal{L}(\theta_1)] - 2\mathbb{E}[\mathcal{L}(\theta_{t+1})]}{\eta(2-\eta \beta)T} +  \frac{\eta \beta\sigma^2}{(2-\eta \beta)}\\
    & \leq \frac{2\mathbb{E}[\mathcal{L}(\theta_1)] - 2\mathbb{E}[\mathcal{L}(\theta^*)]}{\eta(2-\eta \beta)T} +  \frac{\eta \beta\sigma^2}{(2-\eta \beta)}
\end{split}
\end{equation*}

\end{proof}

\section{Proof of Theorem \ref{thm:gap_smooth}}
\label{pf:gap_smooth}
\begin{proof}
We introduce as tools the notion of uniform stability and a corresponding generalization bound from \cite{TrainFaster}.
\begin{myDef}[Uniform Stability]
A randomized learning algorithm $\mathcal{A}$ is $s$-uniformly stable if for all datasets $D, D'$ of size $n$ that differ in one sample, we have
\begin{align*}
    \mathbb{E}[\loss(\theta_D, x) - \loss(\theta_{D'}, x)] \leq s
\end{align*}
for all sample $x$, where $\theta_D$ and $\theta_{D'}$ are respectively the final parameter learned from $D$ and $D'$ with $\mathcal{A}$ and the expectation is taken over the randomness of $\mathcal{A}$.
\end{myDef}

\begin{myTheo}[Generalization with Uniform Stability]
Let $\mathcal{A}$ be $s$-uniformly stable. We have
\begin{align*}
|\mathbb{E}[\hat{\loss}(\theta) -\loss(\theta) ]| \leq s,
\end{align*}
where $\theta$ is the final parameter learned from training data with $\mathcal{A}$ and the expectation is taken over both the randomness of $\mathcal{A}$ and the draw of training data.
\label{thm:GapUniStab}
\end{myTheo}

We can now focus on bounding the uniform stability of DP-SGD, which can be then directly transferred into a bound for generalization gaps with Theorem \ref{thm:GapUniStab}.

Let $D=\{x_1,\ldots,x_N\}, D'=\{x'_1,\ldots,x'_N\}$ be two datasets of size $n$ that differ in one sample(without loss of generality, we assume $x_1\neq x'_1$)

Firstly, we analyze DP-SGD given realizations of the following randomness: Let $\theta_0=\theta'_0$ be a realization of the random initialization of DP-SGD, $\{n_1, \ldots, n_T\} = \{n'_1, \ldots, n'_T\}$ be a realization of Gaussian noise from $\mathcal{N}(0, \sigma^2 C^2 \Id)$ for all $T$ steps of DP-SGD and $\{B_1, \ldots, B_T\}=\{B'_1, \ldots, B'_T\}$ be a realization of indices of samples in all $T$ steps. 

We use $\theta_i$ and $\theta'_i$ to denote respectively the parameters after the $i$-th updates of DP-SGD on $D$ and $D'$, and use $h_i$ to denote $\| \theta_i - \theta'_i \|$. Thus we have $h_0 = 0$. 

In $i$-th step, if the one sample that differs in $D$ and $D'$ is not selected, we have 
\begin{align*}
h_i \leq & h_{i-1} + \frac{\eta}{qn} \cdot \sum_{j\in B_i} \| \text{clip} \left(\nabla \loss_\text{train}(\theta_{i-1}, x_j)\right) \\
&- \text{clip} \left(\nabla \loss_\text{train}(\theta'_{i-1}, x_j)\right) \| \\
\leq & h_{i-1} + \frac{\eta}{qn} |B_i| \cdot \beta h_{i-1} \\
= & h_{i-1} \cdot (1 + \eta \beta)
\end{align*}
where $q$ is the sampling probability(i.e. $qn = |B_i|$ when using fixed-size batches) and the second inequality uses the condition that $\loss_\text{train}$ is $\beta$-smooth(so that $\text{clip} \left( \nabla \loss_\text{train} \right)$ is $\beta$-Lipschitz).

Similarly, if the one sample that differs in $D$ and $D'$ is selected, we have
\begin{align*}
h_i \leq & h_{i-1} + \frac{\eta}{qn} \cdot \sum_{j\in B_i} \| \text{clip} \left(\nabla \loss_\text{train}(\theta_{i-1}, x_j)\right) \\
&- \text{clip} \left(\nabla \loss_\text{train}(\theta'_{i-1}, x'_j)\right) \| \\
\leq & h_{i-1} + \frac{\eta}{qn} (|B_i| - 1) \cdot  \beta h_{i-1} + \frac{\eta}{qn} \cdot 2C \\
\leq & h_{i-1} \cdot (1 + \eta \beta) + \frac{2\eta C}{qn}
\end{align*}

Combining both cases with the inital condition $h_0=0$, we have
\begin{align*}
    h_T \leq N_1 \cdot \frac{2\eta C}{qn} \cdot (1 + \eta \beta)^{T-1},
\end{align*}
where $N_1$ is the total number of steps that contains the different sample(i.e. $x_1$ and $x'_1$).

Since $\loss(\theta, x)$ is $L$-Lipschitz for every $x$, we have
\begin{align*}
    \mathbb{E}[\loss(\theta_T, x) - \loss(\theta'_T, x)] \leq & L \| \theta_T - \theta'_T \|\\
    = & L h_T\\
    \leq & L \cdot N_1 \cdot \frac{2\eta C}{qn} \cdot (1 + \eta \beta)^{T-1}
\end{align*}

By taking expectation on both side over all previously given realizations, we have
\begin{align*}
\mathbb{E}[\loss(\theta_T, x) - \loss(\theta'_T, x)] \leq   & L \cdot \mathbb{E}[N_1] \cdot \frac{2\eta C}{qn} \cdot (1 + \eta \beta)^{T-1}\\
=&L \cdot qT \cdot \frac{2\eta C}{qn} \cdot (1 + \eta \beta)^{T-1}\\
=&\frac{2\eta CLT}{n} \cdot (1 + \eta \beta)^{T-1}
\end{align*}

Since it holds for all $D, D'$ of size $n$ that differ in one sample, by Theorem \ref{thm:GapUniStab}, we have
\begin{align*}
|\mathbb{E}[\hat{\loss}(\theta) -\loss(\theta) ]| \leq \frac{2\eta CLT}{n} \cdot (1 + \eta \beta)^{T-1}.
\end{align*}

\end{proof}

\section{Experimental Setup: Evaluation of Smoothing for PATE}
\label{sec:setup_PATE}
We evaluate PATE on SVHN\cite{SVHN} benchmark, which consists of 73257 images for training and 26032 images for testing. We preserve 10000 samples from its original test set for evaluating model performance and use the remaining 16032 samples as the source of public data.

\begin{table}[!b]  
\renewcommand{\arraystretch}{1.3} 
\newcommand{\tablewidth}{6} 
\caption{Details of Evaluation of Smoothing for PATE} 
\label{tab:PATE_details}

\centering
\resizebox{\linewidth}{!}{ 
\begin{tabular}{c|c|cccc}
\hline
\bfseries parameter & $\#\text{queries}$ & $\#\text{labeled}$ & $\eps$ &$\sigma_{\text{smooth}}$ & $P_{\text{correct}}$ \\ \hline
\multirow{4}{*}{$\sigma_2=40$} & $4000$ & $1323$ & $7.44$ &$0.03$ & $91.38\%$\\
& $8000$ & $2692$ & $11.08$ &$0.03$ & $91.01\%$\\
& $12000$ & $4128$ & $14.08$ & $0.03$ &$91.23\%$ \\
& $16000$ & $5521$ & $16.75$ &$0.025$ & $91.66\%$\\
\hline
\multirow{4}{*}{$\sigma_2=80$} & $4000$ & $1323$ & $3.78$ &$0.04$ & $81.41\%$ \\
& $8000$ & $2692$ & $5.51$ & $0.04$ & $81.35\%$ \\
& $12000$ & $4128$ & $6.90$ & $0.03$ & $81.61\%$ \\
& $16000$ & $5521$ & $8.10$ &$0.03$ & $81.89\%$\\
\hline
\multirow{4}{*}{$\sigma_2=100$} & $4000$ & $1323$ & $3.15$ &$0.03$ & $69.39\%$\\
& $8000$ & $2692$ & $4.57$ & $0.04$ & $70.54\%$ \\
& $12000$ & $4128$ & $5.70$ & $0.03$ & $70.93\%$ \\
& $16000$ & $5521$ & $6.68$ & $0.03$ & $71.26\%$ \\
\hline
\end{tabular}
}
\end{table}

Following \cite{papernot2018scalable}, we use the labels produced by ensembles of $250$ teachers published by \cite{papernot2016semi}. We use Confident-GNMax Aggregator\cite{papernot2018scalable} to aggregate predictions of teachers. We fix $T=300, \sigma_1=200$ and report test accuracy and privacy budgets corresponding to total number of label queries $\#\text{queries}=4000,8000,12000,16000$ with $\sigma_2=40,80,100$. 
In presenting privacy budgets, we fix $\delta$ to be $10^{-5}$ and present only the corresponding $\eps$ for simplicity. To highlight the utility derived from privacy-preserving labels rather than public data, student models in this section are trained in a supervised manner using public data with privacy-preserving labels.

Smoothing is applied to the learning loss as follows when training student models:
\begin{equation*}
    \calL_{\text{smooth}}(\btheta_\text{student}) = 
    \frac{1}{K} \sum_{j=1}^K\calL(\btheta_\text{student} + \mathcal{N}(0, \sigma_{\text{smooth}}^2 \Id)),
\end{equation*}
where $K$ is number of smoothing samples and $\sigma_{\text{smooth}}$ controls the degree of smoothing.
We set $K=10$ in all settings. We use a batch size of $100$ and a learning rate of $0.01$ that decays to $0.001$ in the middle of training. In each setting, we train the student model for sufficiently long and report the highest test accuracy among training. The architecture of the student model is a CNN inherited from the official tutorial of tensorflow/privacy\myendref{note:tensorflow_privacy}. Table \ref{tab:PATE_details} contains other details of evaluation.

\theendnotes
\runningtitle{\name: Boosting Utility of Differentially Private Deep Learning via Randomized Smoothing}
\end{appendices}

\bibliographystyle{abbrv}
\bibliography{ref}

\end{document}